\pdfoutput=1

\documentclass[11pt]{article}

\usepackage{acl}

\usepackage{times}
\usepackage{latexsym}

\usepackage[T1]{fontenc}

\usepackage[utf8]{inputenc}

\usepackage{microtype}

\usepackage{inconsolata}
\usepackage{todonotes}
\usepackage{rotating}
\usepackage{graphicx}
\usepackage{balance} 
\usepackage{lscape}
\usepackage{amsmath}
\usepackage{amssymb}
\usepackage{amsthm}
\usepackage{float}
\usepackage{multirow}
\usepackage{algorithm}
\usepackage{algpseudocode}
\usepackage{tcolorbox}
\usepackage{booktabs}
\usepackage{pifont}
\usepackage{bbm}
\usepackage[normalem]{ulem}
\usepackage{hyperref}
\usepackage{algorithmicx}
\useunder{\uline}{\ul}{}

\newcommand{\KG}{\ensuremath{\mathcal{G}}}

\newcommand{\postriple}{\ensuremath{\mathcal{T}}}
\newcommand{\negtriple}{\ensuremath{\mathcal{T}^-}}
\algdef{SE}[DOWHILE]{Do}{doWhile}{\algorithmicdo}[1]{\algorithmicwhile\ #1}%

\newtheorem{definition}{Definition}
\newtheorem{theorem}{Theorem}
\newtheorem{example}{Example}
\newtheorem{proposition}{Proposition}
\algnewcommand{\LineComment}[1]{\State \(\triangleright\) #1}
\title{ Predictive Multiplicity of Knowledge Graph Embeddings in Link Prediction }

\author{%
  Yuqicheng Zhu$^{\dagger\ddagger}$, Nico Potyka$^\mathsection$, Mojtaba Nayyeri$^\dagger$,  Bo Xiong$^\dagger$ \\
   \textbf{Yunjie He}$^{\dagger\ddagger}$\textbf{,} \textbf{Evgeny Kharlamov}$^{\ddagger\flat}$\textbf{,} \textbf{Steffen Staab}$^{\dagger\natural}$
   \\
   $^\dagger$University of Stuttgart, $^\ddagger$Bosch Center for AI, \\$^\mathsection$Cardiff University, $^\flat$ University of Oslo, $^\natural$ University of Southampton\\
  \texttt{yuqicheng.zhu@de.bosch.com}\\
}

\begin{document}
\maketitle
\begin{abstract}
Knowledge graph embedding (KGE) models are often used to predict missing links for knowledge graphs (KGs). However, multiple KG embeddings can perform almost equally well for link prediction yet give conflicting predictions for unseen queries. This phenomenon is termed \textit{predictive multiplicity} in the literature.
It poses substantial risks for KGE-based applications in high-stake domains but has been overlooked in KGE research. 
We define predictive multiplicity in link prediction, introduce evaluation metrics and measure predictive multiplicity for representative KGE methods on commonly used benchmark datasets.
Our empirical study reveals significant predictive multiplicity in link prediction, with $8\%$ to $39\%$ testing queries exhibiting conflicting predictions.
We address this issue by leveraging voting methods from social choice theory, significantly mitigating conflicts by $66\%$ to $78\%$ in our experiments. 

\end{abstract}
\section{Introduction}
Knowledge graphs (KGs) store factual knowledge of real-world entities and their relationships in the form of triples $\langle \textit{head entity}, \textit{predicate}, \textit{tail entity}\rangle$. 
KGs allow for logical reasoning and answering of queries. Knowledge graph embeddings (KGE) apply machine-learning methods on KGs to provide extra-logical reasoning capabilities exploiting similarities and analogies over knowledge structures \cite{ji2021survey}.

KGE maps entities and predicates into low-dimensional vectors that preserve semantic and structural information of KGs \cite{hogan2021knowledge}.
The learned embeddings can be applied to downstream tasks like link prediction. Given queries in the form of $\langle \textit{head entity}, \textit{predicate}, \textit{?}\rangle$ or $\langle \textit{?}, \textit{predicate}, \textit{tail entity}\rangle$, candidate entities are ranked based on predictive scores provided by KGE models. The positive triples are expected to be ranked higher than the negative triples.

\begin{figure}[t!]
\centering
\includegraphics[width=.48\textwidth]{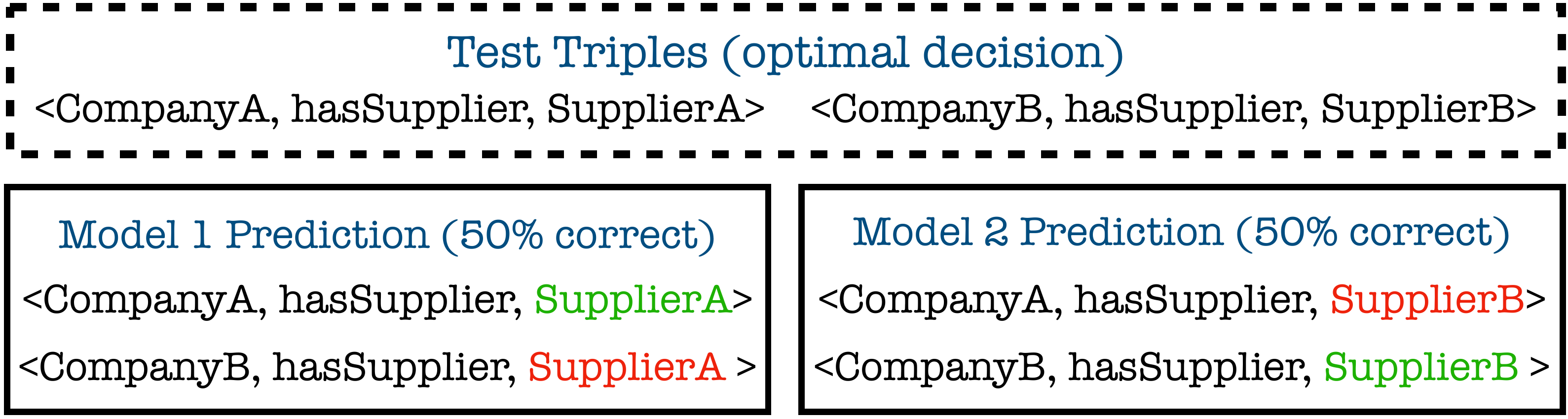}
\caption{An illustration of predictive multiplicity in link prediction lies within the realm of supplier selection for Company A, where model 1 and 2 are trained with the same KGE algorithm (e.g. TransE) but different random seeds. }\label{fig:mp_issue}
\end{figure}

The training of the KG embedding introduces randomness into the resulting model. Sources of randomness include randomized parameter initialization, randomized sequences of positive samples, and randomized negative sampling. 
Given the non-convexity of the training problem, the same KG may lead to various KG embeddings because of the convergence of the training in different local minima. 
While learned embeddings may exhibit comparable performance in link prediction, they may suggest conflicting predictions for an individual query. This phenomenon is referred to as \textit{predictive multiplicity} in recent literature \cite{Marx2020pm,Daniels2023pm, mmsurvey2022}, it is also known as "\textit{Rashomon effect}" and \textit{model multiplicity} in earlier studies \cite{breiman2001statistical}. 
As an example of predictive multiplicity in link prediction, Figure \ref{fig:mp_issue} shows the results of two models that both have an overall accuracy of 50\%, but predict entirely different facts as top 1 recommendation.

Conflicting predictions introduce considerable risks when applying KGE methods in high-stake domains such as medicine or finance. For example, they would affect treatment decisions, affecting patient health outcomes in the context of medical recommendation \cite{gong2021smr}, or switch compounds for confirmatory experiments in drug discovery \cite{mohamed2020discovering}, potentially altering research direction and efficiency. 
Moreover, predictive multiplicity complicates the justification of decisions made from equally accurate models \cite{mmsurvey2022}. For example, when equally accurate models provide contradictory decisions regarding the approval of a loan application \cite{alam2022loan}, the random selection of a model fails to properly justify the ultimate individual decision. Despite its relevance, predictive multiplicity has been overlooked in KGE research.


To the best of our knowledge, this is the first work to study predictive multiplicity for KGE-based link prediction. Our contribution is two-fold: First, we formally define predictive multiplicity in the context of link prediction. Two metrics, \textit{ambiguity} and \textit{discrepancy}, are introduced to measure predictive multiplicity, with an upper bound derived for discrepancy. Evaluating the predictive multiplicity for six representative KGE methods on commonly used benchmark datasets, we observe significant predictive multiplicity behavior in link prediction, with conflicting predictions ranging from $8\%$ to $39\%$ for testing queries.

To address this issue, our second contribution is to investigate the effectiveness of voting methods from social choice theory in mitigating predictive multiplicity in link prediction.
Applying voting methods to aggregate individual rankings yields a more robust ranking that optimizes the collective preference. Our empirical findings demonstrate significant alleviation of predictive multiplicity through voting methods, with the most effective approach reducing conflicting predictions by $66\%$ to $78\%$ for testing queries.

\section{Related Work}
Although prior studies demonstrate the effectiveness of KGE methods in learning complex patterns in KGs \cite{bordes2013translating, sun2019rotate, nickel2011rescal, yang2015distmult, trouillon2016complex, dettmers2018convolutional,DBLP:conf/kdd/XiongZNXP0S22}, fewer works focus on their reliability. Some studies address uncertainty quantification in KGE methods \cite{Safavi2020calibration, Tabacof2020calibration, zhu2024conformalized}, and \citet{peru2021adversarial, peru2021poisoning} explore their adversarial robustness. However, the phenomenon of predictive multiplicity in KGE methods has been overlooked.

The term \textit{model multiplicity} was first discussed in \citet{breiman2001statistical} with the term "\textit{Rashomon Effect}" referring specifically to the phenomenon where there are different weights learned for linear regression with the same error rate. The term \emph{predictive multiplicity} was first introduced by \citet{Marx2020pm}, who explored this behavior in binary classification. 
\citet{Marx2020pm} further investigate predictive multiplicity in probabilistic classification. 
Recent studies also provide evidence of predictive multiplicity for deep models
\cite{black2022select, mehrer2020individual}. We initiate an exploration into the predictive multiplicity behavior within the context of KGE-based link prediction.

While predictive multiplicity offers flexibility in model selection without sacrificing accuracy, diverging predictions can result in unjustifiable final choices. 
\citet{black2022select} propose a method to provide consistent predictions. Given diverging predictions, they first filter them through a specified confidence threshold and select the final prediction using a majority vote. Besides classification problems, predictive multiplicity is also frequently studied for counterfactual explanations \cite{pawelczyk2020counterfactual, jiang2024ensemble}. 

Voting methods can also be seen as ensemble methods.
Ensemble strategies are employed in KGE methods \cite{joshi2022ensemble, xu2021ensemble} during the training phase to increase the model performance. \citet{joshi2022ensemble} focuses on enhancing the accuracy of the triple classification task by aggregating predictions from models trained using different KGE algorithms. \citet{xu2021ensemble} demonstrate that combining multiple low-dimensional models can outperform a single high-dimensional model. However, our approach aggregates rankings using social choice theory in testing time, aiming to alleviate predictive multiplicity by providing more robust rankings. 

\section{Notations and Preliminaries}
\subsection{Knowledge Graph Embedding}
We consider a KG $\KG \subseteq E \times R \times E$ defined over a set $E$
of entities and a set $R$ of relations. The elements in $\KG$ are 
called triples and denoted as $<h,r,t>$.
A KGE model $M_\theta:E\times R\times E\rightarrow\mathbb{R}$ allocates each triple with a predictive score that measures the plausibility that the triple holds 
\cite{bordes2013translating}. The parameters $\theta$ are learned to let $M_\theta$ assign higher predictive scores to positive triples (real facts) while assigning lower predictive scores to negative triples (false facts). This can be achieved for example by minimizing \emph{margin-based ranking loss} \cite{bordes2013translating}:
\begin{equation}
    \mathcal{L}=\sum_{tr\in\postriple}\sum_{tr^-\in\negtriple}\max(0,\gamma-M_\theta(tr)+M_\theta(tr^-)),
\end{equation}
or \emph{cross-entropy loss} \cite{trouillon2016complex}:
\begin{equation}
    \mathcal{L}=\sum_{tr\in\postriple\cup\negtriple}\log(1+\exp(-y_{tr}\cdot M_\theta(tr))),
\end{equation}
where $\gamma$ is a margin hyperparameter, $tr$ refers to a triple $\langle h,r,t\rangle$, $\postriple, \negtriple$ are the sets of positive and negative triples, respectively. The label of a triple, denoted as $y_{tr}$, takes values from the set $\{-1, 1\}$. Here, $y_{tr} = 1$ indicates the triple as positive, while $y_{tr} = -1$ indicates that the triple is negative. The negative triples are typically generated by randomly replacing the head entity or the tail entity in a positive triple with a random entity sampled from the entity set.

\subsection{Social Choice Theory}\label{social choice}
Social choice theory 
studies 
collective decision-making processes, where individual preferences are aggregated to determine a group's overall preference \cite{socialchoice2016}. In this section, we recall some 
basics of social choice theory from \citet{socialchoice2009}. 

We consider a finite set of candidates $C=\{c_1,\dots, c_m\}$ and a finite set of voters $V=\{1,\dots, n\}$, who have different preferences on candidates in $C$. We
represent preferences by a linear order $\succeq$ and let
\begin{itemize}
    \item $c_1\succ c_2 \ \textit{iff}\ c_1\succeq c_2\wedge c_2\not\succeq c_1$ (\emph{strict preference})
    \item $c_1\sim c_2 \ \textit{iff}\ c_1\succeq c_2\wedge c_2\succeq c_1$ (\emph{indifference})
\end{itemize}

We let $\succeq_i$ denote the preference ordering of the $i$-th voter. A  \emph{preference profile} $p: [\succeq_1, \dots, \succeq_n]$ is a list of preference orderings. 
Next, we introduce some interesting voting methods from social choice theory \cite{socialchoice2016}. 

\begin{definition}[Scoring Rule]
    A score vector is a vector $\textbf{w} \in \mathbb{R}^m$ such that $w_1\geq w_2\geq\dots\geq w_m$ and $w_1 > w_m$. Any score vector induces a scoring rule, in which each voter awards $w_1$ points to the top-ranked candidate, $w_2$ points to the second-ranked, and so on. The candidate with the highest total sum of scores wins. 
\end{definition}

\begin{definition}[Majority Voting]
    Majority voting is a scoring rule with the score vector $(1,0,\dots,0)$.
\end{definition}

\begin{definition}[Borda Voting]
    Given $m$ candidates, Borda voting is a scoring rule with the score vector $(m-1,m-2,\dots,0)$.
\end{definition}


\begin{definition}[Range Voting \cite{smith2000range}]
    Given $m$ candidates, range voting is a scoring rule with a score vector $\textbf{w}\in [-1,1]^m$. 
\end{definition}

Additionally, we introduce several properties desirable for the link prediction task in Appendix \ref{app:social}.

\section{Predictive Multiplicity in Link Prediction}
\subsection{Link Prediction}
A query $q\in Q$ is of the form $\langle h,r,?\rangle$ or $\langle ?,r,t\rangle$. We let $tr(q,e)$ denote the corresponding triple $\langle h,r,e\rangle$ or $\langle e,r,t\rangle$, respectively. A KGE model $M_\theta$ can be used to rank the candidate entities for query $q$. We define the ranking $\succeq_{M_\theta, q}$ by
$e_1 \succeq_{M_{\theta, q}} e_2$ iff $M_{\theta, q}(tr(q,e_1)) \geq M_{\theta, q}(tr(q,e_2))$.
We let $R_{\succeq_{M_\theta, q}}(e)$ denote the rank position of a specific candidate entity $e\in E$, that is
\begin{equation}
    R_{\succeq_{M_\theta, q}}(e) = 1 + |\{d \in E \mid d \succeq_{M_\theta, q} e\}|
\end{equation}
 Then the link prediction task can be formulated as a binary classification problem: determine whether a triple is ranked within the top-K predictions:
\begin{equation}
    T_K(M_\theta, tr(q,e)) = \mathbbm{1}[R_{\succeq_{M_\theta, q}}(e)\leq K].
\end{equation}

The performance of link prediction is commonly evaluated by $Hits@K$.
The test set $\mathcal{T}$ contains testing queries $(q, e)$ 
consisting of a query $q$ and a correct answer $e$.
We define the $Hits@K$ function $H_K$ of a KGE model $M_\theta$ as 
\begin{equation}
    \small H_K(M_\theta) = \frac{1}{|\mathcal{T}|}\sum_{(q,e)\in \mathcal{T}}\mathbbm{1}[R_{\succeq_{M_\theta, q}}(e)\leq K]
\end{equation}

\subsection{Definition of Predictive Multiplicity}

We study KGE models that perform similarly in link prediction task in terms of $Hits@K$, i.e. competing models. Following \citet{Marx2020pm}, we will now define a \textit{$\epsilon$-level set} for similar performing models and $\epsilon$ as the \textit{error tolerance}. 

We let $\mathcal{M}$ denote a hypothesis class
of KGE models.
A \emph{baseline model} $M_\theta^*\in\mathcal{M}$ is the KGE model that achieves the highest $Hits@K$ on the validation dataset throughout the hyperparameter optimization process. $D(M_\theta,M_\theta^*)$ measures the difference between baseline model and a competing model with respect to Hits@K. 
\begin{equation}
    D(M_\theta,M_\theta^*) = H_K(M_\theta^*) - H_K(M_\theta).
\end{equation}

\begin{definition}[$\epsilon$-level set]\label{def:epsilon-set}
    Given a baseline KGE model $M_{\theta}^*$ and a hypothesis class $\mathcal{M}$, the $\epsilon$-level set around $M_{\theta}^*$ is the set of all models $M_\theta\in\mathcal{M}$ with a performance difference at most $\epsilon$ in the link prediction task.
    \begin{equation}
        S_\epsilon(M_{\theta}^*):=\{M_\theta\in\mathcal{M}\mid D(M_\theta, M_{\theta}^*)\leq\epsilon\},
    \end{equation}
\end{definition}
Given a testing query set $\mathcal{T}$, predictive multiplicity is defined for testing queries $\tau=(q,e)$ that receive conflicting predictions from competing models.
\begin{definition}[Predictive Multiplicity]\label{def:pm}
    Given a baseline KGE model $M_\theta^*$, an error tolerance $\epsilon$ and a testing query set $\mathcal{T}$, link prediction problem exhibits predictive multiplicity over the $\epsilon$-level set $S_\epsilon(M_\theta^*)$ if there exists a model $M_\theta\in S_\epsilon(M_\theta^*)$ such that $T_K(M_\theta, tr(\tau_i))\not = T_K(M_\theta^*, tr(\tau_i))$ for some $\tau_i\in\mathcal{T}$.
\end{definition}

\subsection{Measuring Predictive Multiplicity}\label{sec:PM_measure}
\textit{Ambiguity} and \textit{discrepancy} are two measures that have been used to quantify predictive multiplicity in classification tasks \cite{Marx2020pm, Daniels2023pm}. We next define them for link prediction.

To make the notation more concise, we use $\Delta(M_\theta, \tau)$ to denote whether a competing model $M_\theta$ provides conflicting predictions compared to the baseline model $M_\theta^*$ for a testing query $\tau = (q,e)$.
\begin{equation}
    \small\Delta(M_\theta, \tau) = 
    \mathbbm{1}[T_K(M_\theta, tr(\tau))\not = T_K(M_\theta^*, tr(\tau))]
\end{equation}

\begin{definition}[Ambiguity]
    Given a testing query set $\mathcal{T}$, the ambiguity of link prediction over the $\epsilon$-level set $S_\epsilon(M_{\theta}^*)$ is the proportion of testing queries that obtain a different prediction by a competing model $M_\theta\in S_\epsilon(M_{\theta}^*)$:
    \begin{equation}
        \alpha_\epsilon(M_{\theta}^*):=\frac{1}{|\mathcal{T}|}\sum_{\tau \in \mathcal{T}}\max_{M_\theta\in S_\epsilon(M_{\theta}^*)}\Delta(M_\theta, \tau)
    \end{equation}
\end{definition}

\begin{definition}[Discrepancy]
    The discrepancy of link prediction over the $\epsilon$-level set $S_\epsilon(M_{\theta}^*)$ is the maximum percentual disagreement
    between the baseline model and a
    competing model $M_\theta\in S_\epsilon(M_{\theta}^*)$:
    \begin{equation}
        \delta_\epsilon(M_{\theta}^*):=\max_{M_\theta\in S_\epsilon(M_{\theta}^*)}\frac{1}{|\mathcal{T}|}\sum_{\tau \in \mathcal{T}}\Delta(M_\theta, \tau)
    \end{equation}
\end{definition}

Ambiguity measures the proportion of testing queries that exhibit predictive multiplicity, while discrepancy captures the largest fraction of test queries for which the predicted answers vary upon switching the baseline model with a competing model. 

\subsection{Bound on Predictive Multiplicity}\label{sec:theory_bound}
In Proposition \ref{pop:bound}, we bound the number of queries with conflicting predictions between the baseline model and a competing model in the $\epsilon$-level set. We provide a proof in Appendix \ref{app:proof}.
\begin{proposition}[Bound on Discrepancy]\label{pop:bound}
    The discrepancy between the baseline model $M_\theta^*$ and any competing model $M_\theta\in S_\epsilon(M_\theta^*)$ obeys:
    \begin{equation}
        \delta_\epsilon(M_\theta^*) \leq 2\cdot (1-H_K(M_\theta^*))+\epsilon
    \end{equation}
\end{proposition}
 The upper bound illustrates how the extent of predictive multiplicity depends on $Hits@K$ of the baseline model. Specifically, a less accurate baseline model theoretically provides greater potential for predictive multiplicity. 
 
\section{Alleviating Predictive Multiplicity using Social Choice Theory}


The predictive multiplicity can be alleviated by improving the robustness of the rankings. Here, robustness means models with similar performance should also provide similar rankings for testing queries. Social choice theory provides a theoretical framework for aggregating individual preferences to determine a group's overall preference \cite{socialchoice2016}. Voting methods from social choice theory can help "smooth out" the randomness in rankings by aggregating individual models \cite{potyka2024robustllm}. 
Intuitively, the candidate entities that are constantly ranked high for all models should also be ranked high in final rankings. 

We next describe ranking aggregation using voting methods with a running example and adapt range voting \cite{smith2000range} to aggregate the predictive scores for the final ranking. 

\subsection{Ranking Aggregation using Voting Methods}\label{sec:aggregation}
For link prediction, given a query $q$ and a KGE model $M_\theta$, the ranking of candidate entities for a query is denoted as $\succeq_{M_\theta,q}$. 
By training KGE models 
with $N$ different random seeds, we obtain a profile for each query $p_q=[\succeq_{M_\theta,q}^1,\dots,\succeq_{M_\theta,q}^N]$.
A ranking aggregation process takes $p_q$ as input and outputs a single ranking. 

We illustrate how to aggregate rankings with voting methods in link prediction task with the following running example.
\begin{example}\label{exp:vote}
    Assume there are in total four entities $\{A,B,C,D\}$ and one relation $r$ in our KG. Given a query $\langle A, r,? \rangle$, three models $[M_\theta^1,M_\theta^2,M_\theta^3]$ sampled from $\epsilon$-level set $S_\epsilon(M_\theta^*)$ provide different rankings in Table \ref{tab:vote_exp}. The predictive scores for candidate entities is shown in brackets after each entity. 
\end{example}

\begin{table}[h!]
    \centering
    \resizebox{.43\textwidth}{!}{
    \begin{tabular}{cc}
    \toprule[1.2pt]
         \textbf{Model ID} & \textbf{Rankings} \\\midrule
         1 & $C (100) \succ_1 B (8) \succ_1 D (6) \succ_1 A (1)$\\
         2 & $B (8) \succ_2 D (7)  \succ_2 C (6) \succ_2 A (5)$\\
         3 & $B (40) \succ_3 C (10) \succ_3 A (2) \succ_3 D (1)$\\ \bottomrule[1.2pt]
    \end{tabular}
    }
    \caption{Rankings of models with corresponding predictive scores for query $\langle A, r,? \rangle$.}
    \label{tab:vote_exp}
\end{table}

We apply all voting methods described in section \ref{social choice} for ranking aggregation. Majority voting and Borda voting aggregate rankings are based on ordinal positions of candidates, while range voting assigns more informative scores to candidates. To adapt range voting in link prediction, we transform the predictive scores into scores within range $[-1,1]$. Concretely, we denote the predictive scores of candidate entities for a query as $\Gamma=[\gamma_1,\dots,\gamma_{|E|}]$ and the score vector of range voting as $\textbf{w}=[w_1,\dots,w_{|E|}]$. We then obtain the score vector based on predictive scores as follows:

\begin{equation}
w_i = 2\times\frac{\gamma_i-\min(\Gamma)}{\max(\Gamma)-\min(\Gamma)}-1.
\end{equation}

Table \ref{tab:vote_scores} shows the scores assigned to candidate entities by voting methods, which are then used to re-rank the entities based on the sum of their voting scores. The resulting aggregated rankings are presented in Table \ref{tab:vote_result}.

\begin{table}[h!]
    \centering
    \resizebox{.45\textwidth}{!}{
    \begin{tabular}{c|cccc|cccc|cccc}
    \toprule[1.2pt]
        \multirow{2}{*}{\textbf{Entity}} & \multicolumn{4}{c}{\textbf{Majority Vote}} & \multicolumn{4}{c}{\textbf{Borda Vote}} & \multicolumn{4}{c}{\textbf{Range Vote}} \\
          & $\succ_1$ & $\succ_2$ & $\succ_3$ & $sum$ & $\succ_1$ & $\succ_2$ & $\succ_3$ & $sum$ & $\succ_1$ & $\succ_2$ & $\succ_3$ & $sum$  \\\midrule
         A & 0 & 0 & 0 & 0  & 0 & 0 & 1 & 1  & -1   & -1    & -0.95 & -2.95  \\
         B & 0 & 1 & 1 & 2  & 2 & 3 & 3 & 8  & -0.85 & 1     & 1 & 1.15  \\
         C & 1 & 0 & 0 & 1  & 3 & 1 & 2 & 6  & 1    & 0.33  & -0.54 & 0.79  \\
         D & 0 & 0 & 0 & 0  & 1 & 2 & 0 & 3  & -0.90 & -0.33 & -1 & -2.23  \\\bottomrule[1.2pt]
    \end{tabular}
    }
    \caption{Ranking aggregation process for Example \ref{exp:vote}.}
    \label{tab:vote_scores}
\end{table}

\begin{table}[h!]
    \centering
    \resizebox{.43\textwidth}{!}{
    \begin{tabular}{cc}
    \toprule[1.2pt]
         \textbf{Voting Method} & \textbf{Rankings} \\\midrule
         Majority Vote & $B (2)  \succ C (1)  \succ D (0) \sim A (0)$\\
         Borda Vote & $B (8) \succ C (6)  \succ D (3) \succ A (1) $\\
         Range Vote & $B (1.15) \succ C (0.79) \succ D (-2.23) \succ A (-2.95) $\\
         \bottomrule[1.2pt]
    \end{tabular}
    }
    \caption{Aggregated rankings of different voting methods for Example \ref{exp:vote}. }
    \label{tab:vote_result}
\end{table}

\section{Experiments}
In this section, we measure the predictive multiplicity in link prediction and apply voting methods from social choice theory. 
Our goals are 
(i) to measure the predictive multiplicity for the link prediction task; 
(ii) to investigate to which extent voting methods can alleviate predictive multiplicity.



\noindent\textbf{Models and Datasets. }The main experiments are conducted for six representative KGE models 
(TransE \cite{bordes2013translating}, 
RotatE \cite{sun2019rotate}, 
RESCAL \cite{nickel2011rescal}, 
DistMult \cite{yang2015distmult}, 
ComplEx \cite{trouillon2016complex}, and
ConvE \cite{dettmers2018convolutional}) on four public benchmark datasets 
(WN18 \cite{bordes2013translating}, 
WN18RR \cite{dettmers2018convolutional}, 
FB15k \cite{bordes2013translating}, and 
FB15k-237 \cite{toutanova2015observed}).
A small dataset Nations \cite{hoyt2022unified} is additionally used for investigating the change of predictive multiplicity with respect to the error tolerance $\epsilon$.
The statistics of benchmark datasets are summarized in Table \ref{tab:dataset}.

\begin{table}[h!]
\resizebox{.48\textwidth}{!}{%
\begin{tabular}{@{}llllll@{}}
\toprule
          & \#Entity & \#Relation & \#Training & \#Validation & \#Test \\ \midrule
WN18      & 40,943   & 18       & 141,442    & 5,000        & 5,000  \\
WN18RR    & 40,943   & 11       & 86,835     & 3,034        & 3,134  \\
FB15k     & 14,951   & 1,345    & 483,142    & 50,000       & 59,071 \\
FB15k-237 & 14,541   & 237      & 272,115    & 17,535       & 20,466 \\ 
Nations & 14   & 55      & 1,592    & 199       & 201 \\ \bottomrule
\end{tabular}%
}
\caption{Statistics of benchmark datasets for link prediction task.}
\label{tab:dataset}
\end{table}

\noindent\textbf{Experiment Settings.} For training KGE, we use the implementation of LibKGE \cite{libkge}. 
All experiments were conducted on a Linux machine with a 40GB NVIDIA A100 SXM4 GPU.

\subsection{Evaluating Predictive Multiplicity}\label{sec:measure_PM}
The $\epsilon$-level set, 
as defined in Definition \ref{def:epsilon-set},
is too large to be evaluated in practice. 
As usual, we will use empirical notions of ambiguity and discrepancy
that are based on a sample of the $\epsilon$-level set that we denote by $S_\epsilon(M_\theta^*)'$.


\noindent\textbf{Constructing the Subset of $\epsilon$-level Set.} 
To construct $S_\epsilon(M_\theta^*)'$, we first obtain the baseline model $M_\theta^*$ by performing 60 trials of hyperparameter search using the strategy in \citet{ruffinelli2019you} (more details in Appendix \ref{app:hyperparameter}) and set $\epsilon$ to $0.01$ (a commonly used value in the literature \cite{Marx2020pm, Daniels2023pm}).
Subsequently, we train a potential competing model using the training configurations of the baseline model with a different random seed.
If the performance difference between the potential competing model and the baseline model is less than $\epsilon$, we add it in $S_\epsilon(M_\theta^*)'$. 
Due to computational constraints, we limit the size of $S_\epsilon(M_\theta^*)'$ to 10 in our experiment. Refer to Algorithm \ref{algo:competing_model} for a pseudocode outlining this process.

\begin{algorithm}[h!]
\caption{Pseudocode for $S_\epsilon(M_\theta^*)'$ construction.}\label{algo:competing_model}
\label{code:data}
\begin{algorithmic}[1]
    \State $M_\theta^*\gets$ Bayesian Optimization for $60$ trials.
    \State $\epsilon\gets 0.01$. \\
    \State $S_{\epsilon}(M_\theta^*)'\gets$ An empty set.
    \While{$|S_{\epsilon}(M_\theta^*)'|\leq 10$}
        \State $M_\theta\gets$ Retrain $M_\theta^*$ with a different random seed.
        \If{$D(M_\theta, M_\theta^*)\leq\epsilon$}
            \State $S_{\epsilon}(M_\theta^*)'$ add $M_\theta$.
        \EndIf
    \EndWhile
    \State \Return $S_{\epsilon}(M_\theta^*)'$
\end{algorithmic}
\end{algorithm}

\noindent\textbf{Evaluation Metrics.} We evaluate the accuracy of link prediction with $Hits@K$ and the predictive multiplicity with ambiguity and discrepancy. 
Note that in our experiment, ambiguity and discrepancy are measured
by their empirical counterpart over the $\epsilon$-level set approximation $S_{\epsilon}(M_\theta^*)'$. To distinguish these metrics from previous definitions in section \ref{sec:PM_measure}, we denote them as $\hat\alpha_\epsilon$ and $\hat\delta_\epsilon$
and call them empirical ambiguity and discrepancy, respectively.

\noindent\textbf{Evaluation Procedure.} We demonstrate the evaluation procedure in Algorithm \ref{algo:evaluation}. 
We denote $Aggregate(A, S_{agg})$ as a procedure to aggregate rankings predicted by models in $S_{agg}$ using a voting method $A$ (detailed in section \ref{sec:aggregation}). 
The result of $Aggregate(A, S_{agg})$ can be viewed as a new KGE model $M_{agg}$ that predicts the aggregated rankings.
train(config($M$), $seed$) denotes the training process of a KGE model, which adopts the same training configurations (including the training graph, hyperparameters, etc.) of a pre-trained model $M$ with a specific random seed. 

\begin{algorithm}[h!]
\caption{Pseudocode for evaluation.}\label{algo:evaluation}
\label{code:evaluation}
\begin{algorithmic}[1]
    \Require $S_{\epsilon}(M_\theta^*)'$
    \State $S\gets$ An empty set. \Comment{Initialize evaluation set}
    \If{not apply voting method $A$}
        \State $S\gets S_{\epsilon}(M_\theta^*)'$
    \Else
        \For{\textbf{each} $M_\theta$ \textbf{in} $S_{\epsilon}(M_\theta^*)'$}
            \State $S_{agg}\gets$ An empty set.
            \For{$i \leftarrow 1$ to $10$}
                \State $seed_i\gets$ generateRandomSeed()
                \State $\hat M_\theta\gets$ train(config($M_\theta$), $seed_i$).
                \State $S_{agg}\gets S_{agg}\cup \{\hat M_\theta\}$
            \EndFor
            \State $M_{agg}\gets$ Aggregate(A, $S_{agg}$).
            \State $S\gets S\cup \{M_{agg}\}$.
        \EndFor
    \EndIf
    \\
    \State Evaluate $Hits@K$ for all models in $S$ and report the average value.
    \State Evaluate $\hat\alpha_\epsilon$ and $\hat\delta_\epsilon$ for $S$.
\end{algorithmic}
\end{algorithm}

For each KGE method and benchmark dataset, we first construct a set of competing models, $S_\epsilon(M_\theta^*)'$. Without employing voting methods, we assess $Hits@K$, $\hat\alpha_\epsilon$, and $\hat\delta_\epsilon$ over $S_\epsilon(M_\theta^*)'$. Otherwise, we collect a set of models $S_{agg}$ for each model $M_\theta$ in $S_\epsilon(M_\theta^*)'$ by training 10 models using the configurations of $M_\theta$ with different random seeds. Subsequently, we aggregate the models in $S_{agg}$ with a voting method $A$ to get an "aggregated" model $M_{agg}$ for each $M_\theta$, and then measure all metrics over the set of aggregated models.  

\begin{table}[h!]
\centering
\resizebox{0.48\textwidth}{!}{%
\begin{tabular}{cc|ccc|cccc}
\toprule[2pt]
\multirow{2}{*}{Models}& \multirow{2}{*}{Baselines} & \multicolumn{3}{c}{WN18RR} & \multicolumn{3}{c}{FB15k237}\\
 & & $Hits@10\uparrow$ & $\hat\alpha_\epsilon\downarrow$ & $\hat\delta_\epsilon\downarrow$ & $Hits@10\uparrow$ & $\hat\alpha_\epsilon\downarrow$ & $\hat\delta_\epsilon\downarrow$ \\\midrule
\multirow{4}{*}{\rotatebox{90}{TransE}}
& w/o & 0.518 & 0.076 & 0.034  & 0.455 & 0.385 & 0.145 \\
& major & 0.055 & 0.096 & 0.045 & 0.155 & 0.171 & 0.081 \\
& Borda & 0.482 & 0.032 & 0.016 & 0.456 & 0.110 & 0.044 \\
& range & {\ul 0.519} & {\ul \textbf{0.017}} & {\ul \textbf{0.009}} & {\ul 0.470} & {\ul 0.101} & {\ul 0.041} \\\midrule
 
\multirow{4}{*}{\rotatebox{90}{RotatE}}
 & w/o & 0.547 & 0.195 & 0.074 & 0.520 & 0.163 & 0.064 \\
& major & 0.413 & 0.064 & 0.029 & 0.204 & 0.104 & 0.053 \\
& Borda & 0.564 & 0.062 & 0.028 & 0.523 & 0.039 & 0.017 \\
& range & {\ul \textbf{0.578}} & {\ul 0.051} & {\ul 0.022} & {\ul \textbf{0.524}} & {\ul \textbf{0.037}} & {\ul \textbf{0.016}} \\\midrule
 
\multirow{4}{*}{\rotatebox{90}{RESCAL}} 
   & w/o & 0.517 & 0.248 & 0.095 & 0.482 & 0.375 & 0.140 \\
& major & 0.198 & 0.108 & 0.054 & 0.145 & 0.165 & 0.089 \\
& Borda & 0.561 & 0.099 & 0.043 & 0.485 & 0.107 & 0.048 \\
& range & {\ul 0.575} & {\ul 0.084} & {\ul 0.034} & {\ul 0.498} & {\ul 0.098} & {\ul 0.042} \\\midrule
 
\multirow{4}{*}{\rotatebox{90}{DistMult}} 
   & w/o & 0.526 & 0.169 & 0.068 & 0.476 & 0.320 & 0.120 \\
& major & 0.185 & 0.078 & 0.037 & 0.144 & 0.124 & 0.059 \\
& Borda & 0.524 & 0.055 & 0.024 & 0.475 & 0.088 & 0.037 \\
& range & {\ul 0.542} & {\ul 0.048} & {\ul 0.021} & {\ul 0.488} & {\ul 0.082} & {\ul 0.034} \\\midrule
 
\multirow{4}{*}{\rotatebox{90}{ComplEx}} 
   & w/o & 0.541 & 0.217 & 0.085 & 0.482 & 0.308 & 0.116 \\
& major & 0.243 & 0.243 & 0.126 & 0.145 & 0.121 & 0.055 \\
& Borda & 0.559 & 0.067 & 0.030 & 0.480 & 0.087 & 0.036 \\
& range & {\ul 0.573} & {\ul 0.058} & {\ul 0.024} & {\ul 0.493} & {\ul 0.082} & {\ul 0.032} \\\midrule
 
\multirow{4}{*}{\rotatebox{90}{ConvE}} 
   & w/o & 0.500 & 0.222 & 0.088 & 0.474 & 0.340 & 0.130 \\
& major & 0.185 & 0.092 & 0.047 & 0.150 & 0.154 & 0.074 \\
& Borda & 0.522 & 0.082 & 0.035 & 0.474 & 0.092 & 0.039 \\
& range & {\ul 0.534} & {\ul 0.068} & {\ul 0.027} & {\ul 0.486} & {\ul 0.085} & {\ul 0.034}\\ \bottomrule[2pt]
\end{tabular}
}
\caption{This table compares the accuracy and predictive multiplicity of applying different voting methods on six representative KGE models and two benchmark datasets, WN18RR and FB15k237. We underline the best values for each model-dataset pair and boldface the global optimal values. (Results for more datasets see Table \ref{tab:full_main_result} in Appendix \ref{app:full_results}.)}\label{tab:main_result}
\end{table}

\noindent\textbf{Results.} We present the results of predictive multiplicity of link prediction in Table \ref{tab:main_result}. For benchmark datasets WN18RR, FB15k237 and six KGE representative methods, 
we observe that competing models with less than $1\%$ error tolerance ($\epsilon = 0.01$) assign conflicting predictions for $8\%$ to $39\%$ of testing queries ($\hat\alpha_\epsilon$).
Voting methods effectively mitigate the issue of predictive multiplicity.
Majority voting generally reduces conflicting predictions but also decreases $Hit@K$ substantially.
Borda voting yields comparable $Hits@K$ and significantly alleviate predictive multiplicity.
Range voting consistently outperforms other methods in terms of $Hits@K$ and substantially reduces predictive multiplicity, resulting in a relative decrease of 66\% to 78\% in empirical ambiguity ($\hat\alpha_\epsilon$) and 64\% to 76\% in empirical discrepancy ($\hat\delta_\epsilon$).


We focus on link prediction for recommendation, emphasizing the importance of whether true facts are ranked within the top-K. In Appendix \ref{app:query_answering}, we extend our analysis to link prediction within a query answering context, where the objective is to determine whether competing models yield similar/same answer sets. Comparable conclusions can be drawn within that context as well.


\subsection{Further Analysis}

\subsubsection{Investigating Predictive Multiplicity wrt. Error Tolerance}
\begin{figure}[h!]
\centering
\includegraphics[width=.48\textwidth]{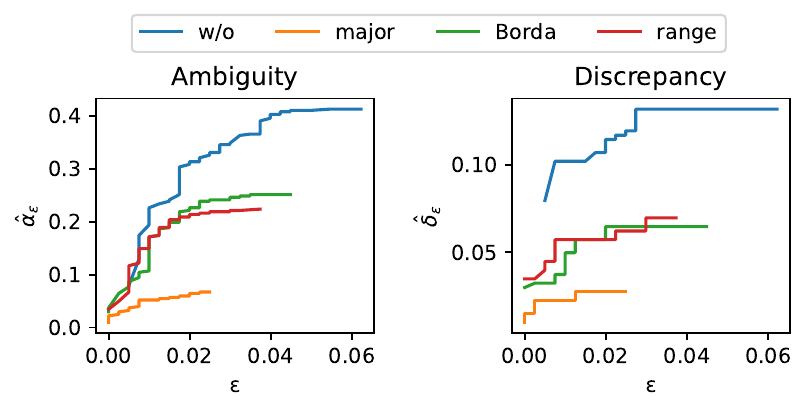}
\caption{Predictive multiplicity for ComplEx on Nations dataset wrt. $\epsilon$. }\label{fig:epsilon_curve}
\end{figure}

We conduct experiment for ComplEx on Nations to investigate the influence of $\epsilon$ on predictive multiplicity. The procedure follows Algorithm \ref{algo:evaluation} with thirty values of $\epsilon$ spanning the range from $0$ to $0.06$. We represent the results in Figure \ref{fig:epsilon_curve}. Our observations confirm the expectation in section \ref{sec:theory_bound}: both predictive multiplicity metrics increase with larger values of $\epsilon$. Employing voting methods consistently reduces both ambiguity and discrepancy across all $\epsilon$ values, with a more pronounced effect observed for larger $\epsilon$. Notably, even at $\epsilon=0$, conflicting predictions persist, underscoring the necessity to report predictive multiplicity even for equally accurate models. Additionally, we observe that the change of $\epsilon$ has negligible effects on $Hits@K$, as detailed in Appendix \ref{app:acc}.

\subsubsection{Investigating the Number of Models for Aggregation}
\begin{figure}[h!]
\centering
\includegraphics[width=.48\textwidth]{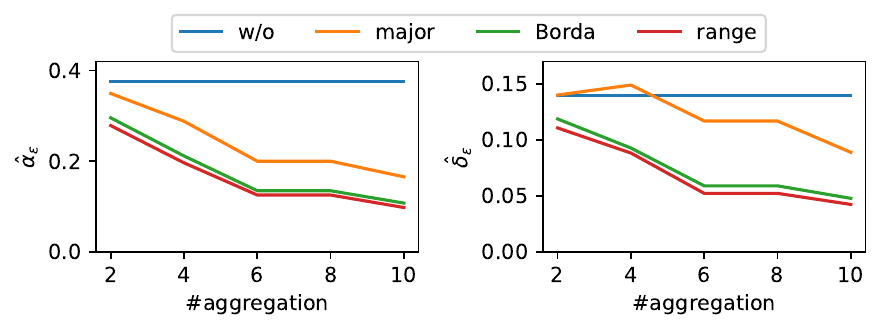}
\caption{Investigation of the predictive multiplicity with respect to the number of models used for voting methods. Due to page limit, we only show the results of RESCAL on FB15k237 in this figure, we put more results in appendix \ref{app:agg_exp}. }\label{fig:agg_exp}
\end{figure}

In Figure \ref{fig:agg_exp}, we investigate the predictive multiplicity metrics in relation to the number of models employed for ranking aggregation. Employing a larger number of models for aggregation yields a more notable alleviation of predictive multiplicity. Remarkably, even with a relatively small number of aggregated models, substantial improvements in predictive multiplicity can be attained. Furthermore, change of the number of models for aggregation does not notably affect $Hits@K$ (Figure \ref{fig:agg_exp_acc_wn18} - \ref{fig:agg_exp_acc_fb15k237} in Appendix \ref{app:agg_exp}).

\subsubsection{Investigating the Predictive Multiplicity wrt. Entity/Relation Frequency}
Most entities/relations only have a few facts in KGs \cite{xiong2018long}. There are more possible embeddings or more uncertainty for those relations/entities since they are less constrained by the existing facts in KG during training. Intuitively, there might be more significant predictive multiplicity behavior for queries containing those entities/relations. 

\begin{table}[h!]
\centering
\resizebox{.45\textwidth}{!}{%
\begin{tabular}{cc|cc|cc}
\toprule[1.5pt]
 &  & \multicolumn{2}{c}{w/o} & \multicolumn{2}{c}{range vote} \\
Var.1 & Var.2 & $\rho$ & p-value & $\rho$ & p-value \\ \midrule
Rel. Fre & $\hat\alpha_\epsilon$ & -0.349 & <0.001 & -0.156 & <0.001 \\
Rel. Fre & $\hat\delta_\epsilon$ & -0.400 & <0.001 & -0.204 & <0.001 \\
Ent. Fre & $\hat\alpha_\epsilon$ & -0.106 & <0.001 & -0.098 & <0.001 \\
Ent. Fre & $\hat\delta_\epsilon$ & -0.114 & <0.001 & -0.103 & <0.001 \\
\bottomrule[1.5pt]
\end{tabular}%
}
\caption{This table presents the correlation between entity/relation frequency and $\hat\alpha_\epsilon$ and $\hat\delta_\epsilon$, with Spearman's coefficient ($\rho$) and its p-value. Columns 3 and 4 show results without applying voting method, while columns 5 and 6 show  results with range voting. }
\label{tab:spearman}
\end{table}

We conduct hypothesis tests using Spearman's coefficient ($\rho$) to assess the correlation between entity/relation frequency (i.e., the number of triples containing the target entity/relation) and predictive multiplicity metrics ($\hat\alpha_\epsilon$ and $\hat\delta_\epsilon$). $\rho$ ranges from -1 to 1, indicating the strength and direction of the correlation: close to 1 implies a positive monotonic relationship, while close to -1 implies a monotonic negative relationship.

We count entity/relation frequencies (Ent. Fre and Rel. Fre) as variable 1 and calculate $\hat\alpha_\epsilon$ and $\hat\delta_\epsilon$ for six KGE methods on entity/relation- specific subsets of all datasets as variable 2. Results in Table \ref{tab:spearman} show a significant negative correlation, confirming our conjecture. Notably, applying range voting weakens this correlation, potentially due to its effectiveness in alleviating predictive multiplicity for queries with higher uncertainty.

\section{Discussing Other Influential Factors of Predictive Multiplicity}\label{sec:influence}
In this section, we discuss additional factors 
that
may influence predictive multiplicity, namely expressiveness and inference patterns. 
We briefly introduce these two notions and then discuss
some observations regarding their relationship to predictive
multiplicity.

\noindent\textbf{Expressiveness.} 
The expressiveness of KGE models refers to the ability of modeling an arbitrary KG.
Following \cite{pavlovic2023expressive, wang2018multi},
we call a KGE model fully expressive if we can find a parameter set such that the model predicts all training triples correctly. 
Intuitively, more expressive models
can represent
more possible embeddings that fit the training graph, thereby allowing more "room" for multiplicity. 

\noindent\textbf{Inference Patterns.} 
Inference patterns refer to the logic rules used to derive new triples from the observed facts in KGs. The generalization capabilities of KGE is usually analysed based on inference patterns that KGE model can capture \cite{abbound2020boxe}. 
For instance, TransE can capture inverse patterns, wherein $r_1(X,Y)$ implies $r_2(Y,X)$, suggesting that the testing triple $\langle e_1, r_2, e_2\rangle$ can be correctly predicted with low uncertainty if $\langle e_2, r_1, e_1\rangle$ is present in the training graph.
Theoretically, if the KGE method effectively captures the inference patterns for the testing triple, we would expect fewer conflicts from competing models.

\begin{figure}[h!]
\centering
\includegraphics[width=.48\textwidth]{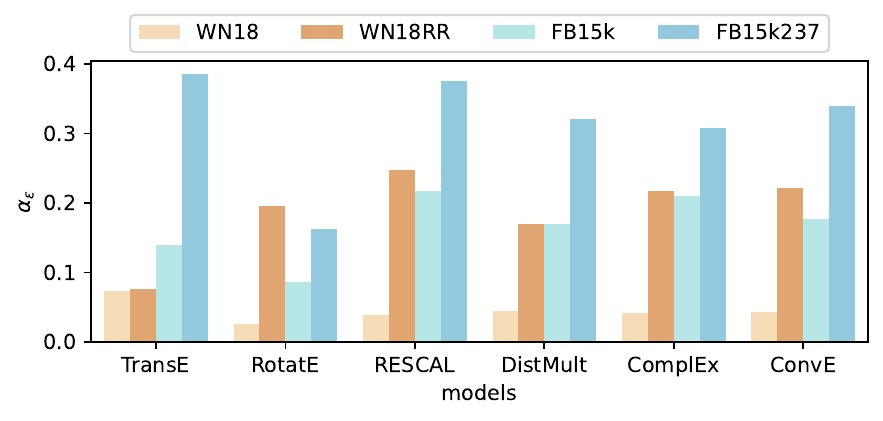}
\caption{We demonstrate the ambiguity for 10 competing models on WN18, WN18RR, FB15k and FB15k237 in this figure. }\label{fig:inference}
\end{figure}

\noindent\textbf{Observations.} 
According to \citet[Table 1]{wang2018multi}, RESCAL and ComplEx are more expressive than DistMult when considering similar embedding dimensions. We observe that RESCAL and ComplEx associate with larger values of ambiguity and discrepancy than DistMult in Table \ref{tab:main_result}, aligning with our conjecture regarding expressiveness. 
Furthermore, WN18 and FB15k are known to suffer from test leakage due to inverse relations \cite{toutanova2015observed}, meaning that many test triples can be easily derived by the inverse pattern. WN18RR and FB15k-237 delete inverse relations to address this issue \cite{toutanova2015observed, dettmers2018convolutional}. In Figure \ref{fig:inference}, we note a consistent trend where competing models exhibit fewer conflicting predictions on WN18 and FB15k compared to WN18RR and FB15k237. This observation supports our conjecture regarding inference patterns, as the absence of even a single inference pattern notably increase the number of conflicting predictions.

The result of TransE on FB15k237 appears to be an outlier, marked by its low expressiveness but the highest ambiguity and discrepancy among all KGE methods.
However, in FB15k237, numerous training triples involve symmetric relations, with testing triples inferrable through symmetric patterns \cite{rim2021behavioral}. Since TransE fails to represent symmetric triple pair ($\langle e_1, r, e_2\rangle$ and $\langle e_2, r, e_1\rangle$) simultaneously and lacks the capability to capture symmetric patterns, it may therefore exhibit additional predictive multiplicity. 

\section{Conclusion}
In this paper, we define and measure the predictive multiplicity in link prediction. 
We measure the predictive multiplicity with empirical ambiguity and discrepancy for representative KGE methods on commonly used benchmark datasets. 
Our empirical study reveals significant predictive multiplicity in link prediction, and we demonstrate the effectiveness of applying voting methods. We also discuss several potential factors that could influence predictive multiplicity in link prediction. 

Furthermore, according to Proposition \ref{pop:bound}, predictive multiplicity depends on the accuracy of the baseline model and error tolerance ($\epsilon$). A less accurate baseline model or larger $\epsilon$ allows for more predictive multiplicity. Given the typically low accuracy in link prediction and the existence of conflicting predictions even when $\epsilon=0$, a considerable number of conflicting predictions may arise from competing models in practice, posing significant risks in safety-critical domains.  
Hence, we advocate for the measurement, reporting, and mitigation of predictive multiplicity in link prediction within these domains.


\section{Limitations}


In Section \ref{sec:influence}, we offer conjectures regarding the relationship between influential factors and predictive multiplicity. Our findings only show that our conjectures are potentially reasonable, but no conclusions can be drawn based on them. A systematic analysis necessitates quantifying expressiveness, inference patterns, which falls outside the scope of our paper but is a promising avenue for future research.

To mitigate predictive multiplicity, employing voting methods derived from social choice theory emerges as a straightforward yet effective strategy. However, voting-based ranking aggregation requires training multiple competing models from scratch, which can be time/computational consuming. Addressing predictive multiplicity during the training phase is considered as next step.
Furthermore, more advanced voting methods such as partial Borda voting \cite{cullinan2014borda} could be explored in the future, which aggregates only partial rankings to reduce memory requirements during the aggregation step.

\section{Ethics Statement}
In this study, we emphasize the importance of reporting and dealing with predictive multiplicity to ensure fair and transparent decision-making processes for KGE-based applications. Failure to account for predictive multiplicity may lead decision-makers to select models that align with their personal preferences, potentially resulting in unfair outcomes for individuals. By neglecting to report predictive multiplicity of KGE models, decision-makers risk undermining the integrity and equity of the decision-making process. 

\section{Acknowledgements}
The authors thank the International Max Planck Research School for Intelligent Systems (IMPRS-IS) for supporting Yuqicheng Zhu, Bo Xiong and Yunjie He. The work was partially supported by the Horizon Europe projects EnrichMyData (Grant Agreement No.101070284), Graph-Massivier (Grant Agreement No.101093202) and Dome 4.0 (Grant Agreement No.953163).

This research has also been partially funded by the Deutsche Forschungsgemeinschaft (DFG, German Research Foundation) – SFB-1574 – 471687386. 
Deutsche Forschungsgemeinschaft (DFG, German Research Foundation) under Germany’s Excellence Strategy - EXC 2075 - 390740016, the Stuttgart Center for Simulation Science (SimTech), the European Union’s Horizon 2020 research and the the Bundesministerium für Wirtschaft und Energie (BMWi), Grant Agreement No. 01MK20008F.

\bibliography{anthology}

\newpage
\appendix

\section{Properties of Voting Methods from Social Choice Theory}
\label{app:social}
All voting methods were proposed to aggregated preferences in an intuitive "fair" way. However, for some cases, they may fail unintendedly. Thus, precisely defined properties - appealing behaviors that the voting methods satisfy, are investigated in social choice theory \cite{socialchoice2016}. 

We introduce some properties from \cite{socialchoice2016} that are desirable for link prediction.  Recall that a \emph{social choice function} is a function $f$ mapping from the set of all possible profiles $\mathcal{P}$ to a non-empty subset of possible candidate $C$. Given a finite set of voters $N=\{1,\dots,n\}$ and a profile $p=[\succeq_1,\dots,\succeq_n]$, $f$ is called:
\begin{itemize}
    \item \textbf{anonymous}: if $f$ does
    not depend on the identity voters, i.e.,  if for every bijective function $\pi: V \rightarrow V$, we have $f([\succeq_1,\dots,\succeq_n])=f([\succeq_{\pi(1)},\dots,\succeq_{\pi(n)}])$
    .
    \item \textbf{neutral}: if $f$ does not depend on the 
    identity of candidates, i.e., if two candidates are exchanged in every preference ordering in $p$, the outcome will change accordingly.
    \item \textbf{Pareto-optimal}: if candidate $c_A$ is ranked higher than candidate $c_B$ in all preference orderings, then $c_B \not \in f(p)$.
    \item \textbf{reinforcing}: If $p_1, p_2$ are disjoint profiles and $f(p_1) \cap f(p_2) \neq \emptyset$ then $f(p_1) \cap f(p_2) = f(p_1 \cup p_2)$.
    \item \textbf{monotonic}: if whenever a profile $p$ is changed to $p'$ by having one voter lifting the winning candidate, $f(p) = f(p')$.
\end{itemize}

\begin{theorem}[\citet{young1975social}]\label{theorem:young}
    Suppose that V is a voting method that requires voters to rank the candidates. Then, V is anonymous, neutral and reinforcing if and only if the method is a scoring rule.
\end{theorem}

According to Theorem \ref{theorem:young}, majority vote and Borda vote as scoring rules are anonymous, neutral and reinforcing.

Note simply averaging the predictive scores does not satisfy some relevant properties for providing such as anonymity. That means KGE models with higher predictive scores for the top ranked entity would dominate the final decision. Therefore, we do not consider averaging as baseline in our paper.

A \textit{social welfare function} $f_w$ is a mapping from the set of all possible profiles $\mathcal{P}$ to a set of all linear orders on $C$. We next introduce some properties of $f_w$.

$f_w$ is:
\begin{itemize}
    \item \textbf{weakly Paretian}: for $c_1, c_2\in C$, if $c_1\prec_i c_2$ for all $i\in N$, then $c_1\prec c_2$.
    \item \textbf{independent of irrelevant alternatives (IIA)}: if for any $c_1, c_2\in C$, the relative ranking of $c_1$ and $c_2$ only depends on the relative rankings of $c_1$ and $c_2$ provided by the voters - but not on how the voters rank some third candidate $c_3$.
    \item a \textbf{dictatorship}: if there exists a voter $i^*\in N$ such that, for all $c_1, c_2\in C$, $c_1\prec_{i^*} c_2$ implies $c_1\prec c_2$.
\end{itemize}
\begin{theorem}[\citet{arrow1951}]\label{theorem:arrow}
    When there are three or more alternatives, then every $f_w$ that is weakly Paretian and IIA must be a dictatorship.
\end{theorem}

Majority vote and Borda vote are both weakly Paretian and non-dictatorship \cite{socialchoice2016}, therefore according to Theorem \ref{theorem:arrow}, they are not IIA. However, range vote as a cardinal voting method meet the Arrow's conditions and additionally provide "maximum information" (i.e. provide their opinion of the maximum possible number of candidates) \cite{vasiljev2014cardinal, smith2000range}.


\section{Proof of Proposition \ref{pop:bound}}\label{app:proof}
\begin{proof}
Given a set of testing queries $\mathcal{T}=\{(q_1,e_1), \dots, (q_n,e_n)\}$,
we let $\hat y \in \mathbb{R}^{n}$,
$y_i = T_K(M^*_\theta, tr(q_i,e_i))$
be the vector that contains a $1$ if 
the baseline model regards $e_i$ as a valid answer. 
Similarly, we let $y' \in \mathbb{R}^{n}$,
$y_i = T_K(M_\theta, tr(q_i,e_i))$
be the corresponding vector for a competing model $M_\theta\in S_\epsilon(M_\theta^*)$. 

Let $\mathbf{1} \in \mathbb{R}^n$ be a vector consisting
only of ones. 
Then we can express the proportion of testing triples not ranked in top-K as $\frac{1}{n}||\mathbf{1}-\hat y||_1$ and $\frac{1}{n}||\mathbf{1}-y'||_1$ for the baseline and competing model, respectively. 
We let $\delta(M_A, M_B)$  denote the discrepancy 
between two models $M_A, M_B\in\mathcal{M}$.
\begin{align*}
    \delta(M_A, M_B)&:=\\
    &\frac{1}{n}\sum_{\tau\in\mathcal{T}}^n\mathbbm{1}[T_K(M_A, \tau)\not = [T_K(M_B, \tau)]
\end{align*}
We can then rewrite 
\begin{align*}
\delta(M_\theta^*, M_\theta)
&= \frac{1}{n}||y'-\hat y||_1 \\
&\leq \frac{1}{n}||\mathbf{1}-y'||_1 + \frac{1}{n}||\mathbf{1}-\hat y||_1 \\
&= (1 - H_K(M_\theta)) + (1-H_K(M_\theta^*)) \\
&\leq 2- H_K(M_\theta^*) + \epsilon -H_K(M_\theta^*), 
\end{align*}
where we used the triangle inequality and symmetry
of the L1-norm for the first
inequality and the definition of $S_\epsilon(M_\theta^*)$ 
 for the second.
Since
$\delta_\epsilon(M_{\theta}^*) =
\max_{M'_\theta\in S_\epsilon(M_{\theta}^*)}\delta(M_\theta^*, M'_\theta)$,
we have
\begin{equation*}
      \delta_{\epsilon}(M_\theta^*)\leq 2\cdot(1-H_K(M_\theta^*))+\epsilon.
\end{equation*}
\end{proof}

\section{More Experiment Settings}\label{app:hyperparameter}
\subsection{Personal Identification Issue in FB15k and FB15k237}
While FB15k and FB15k237 contain information about individuals, it typically focuses on well-known public figures such as celebrities, politicians, and historical figures. Since this information is already widely available online and in various public sources, its inclusion in Freebase doesn't significantly compromise individual privacy compared to datasets containing sensitive personal information.

\subsection{Change of $S_\epsilon(M_\theta)'$ after Applying Voting Methods}
Theoretically, we need to ensure that the aggregated models within the evaluation set $S$ should also have exactly the same $\epsilon$ with the original set of competing models $S_\epsilon(M_\theta^*)'$. In order to do that, the pseudocode of evaluating predictive multiplicity should look like following:

\begin{algorithm}[h!]
\caption{Pseudocode for evaluation (in theory).}\label{algo:evaluation_theory}
\begin{algorithmic}[1]
    \Require $S_{\epsilon}(M_\theta^*)'$
    \State $S\gets$ An empty set. \Comment{Initialize evaluation set}
    \If{not apply voting methods}
        \State $S\gets S_{\epsilon}(M_\theta^*)'$
    \Else
        \LineComment{Aggregation for the baseline model}
        \State $S_{agg}^*\gets$ An empty set.
        \For{$i \leftarrow 1$ to $10$}
            \State $seed_i\gets$ RandomSeed()
            \State $\hat M_\theta^*\gets$ train(conf($M_\theta^*$), $seed_i$).
            \State $S_{agg}^*\gets S_{agg}^*\cup \{\hat M_\theta^*\}$
        \EndFor
        \State $M_{agg}^*\gets$ Aggregate(A, $S_{agg}^*$).
        \State $S\gets S\cup \{M_{agg}^*\}$.\\
        
        \LineComment{Aggregation for the competing models}
        \While{$|S|\leq 10$}
            \State $S_{agg}\gets$ An empty set.
            \Do
                \For{$i \leftarrow 1$ to $10$}
                    \State $seed_i\gets$ RandomSeed()
                    \State $\hat M_\theta\gets$ train(conf($M_\theta^*$), $seed_i$).
                    \State $S_{agg}\gets S_{agg}\cup \hat \{M_\theta\}$
                \EndFor
                \State $M_{agg}\gets$ Aggregate(A, $S_{agg}$).
            \doWhile{$D(M_{agg}^*, M_{agg})\leq\epsilon$}
            \State $S\gets S\cup \{M_{agg}\}$.
        \EndWhile
    \EndIf
    \\
    \State Evaluate $Hits@K$ for all models in $S$ and report the average value.
    \State Evaluate $\hat\alpha_\epsilon$ and $\hat\delta_\epsilon$ for $S$.
\end{algorithmic}
\end{algorithm}

Recall from Algorithm \ref{algo:evaluation}, we denote $Aggregate(A, S_{agg})$ as a procedure to aggregate rankings predicted by models in $S_{agg}$ using a voting method $A$ (detailed in section \ref{sec:aggregation}). 
The result of $Aggregate(A, S_{agg})$ can be viewed as a new KGE model $M_{agg}$ that predicts the aggregated rankings.
train(conf($M$), $seed$) denotes the training process of a KGE model, which adopts the same training configurations (including the training graph, hyperparameters, etc.) of a pre-trained model $M$ with a specific random seed. 

\begin{figure}[h!]
\centering
\includegraphics[width=.48\textwidth]{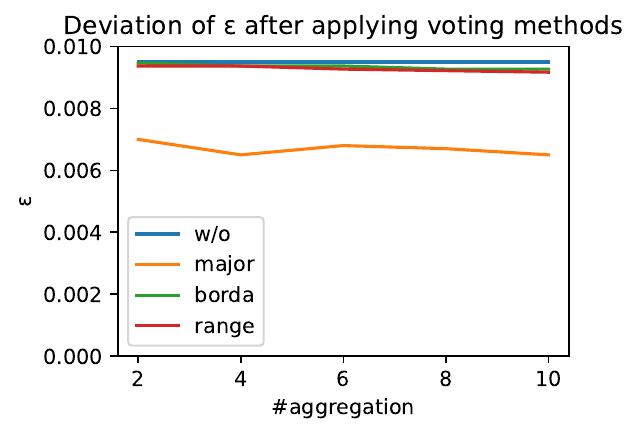}
\caption{Deviation of $\epsilon$ after voting methods wrt. the number of models used for aggregation (results for RESCAL on FB15k237). }\label{fig:eps_dev}
\end{figure}

The procedure described in Algorithm \ref{algo:evaluation} can not guarantee to have same $\epsilon$ for both $S$ and $S_\epsilon(M_\theta^*)'$, since $Hits@K$ changes after applying voting methods. However, obtaining a desirable aggregated model with the do-while loop (from line 19 to line 27) in Algorithm \ref{algo:evaluation_theory} can be very time/computational consuming (approximately 10 hours for each loop). Therefore, we obtain the aggregated model from each competing model in $S_\epsilon(M_\theta^*)'$ to reduce the training effort in Algorithm \ref{algo:evaluation}. Empirically, we observe a negligible deviation of $\epsilon$ after applying the evaluation procedure of Algorithm \ref{algo:evaluation}, see Figure \ref{fig:eps_dev}. This level of $\epsilon$ deviation should not significantly change our claims.

\subsection{Hyperparameter Search}
To get the baseline model $M_\theta^*$, we use PyTorch-based library LibKGE \cite{libkge} (MIT-license) and basically follow the hyperparameter search strategy in \citet{ruffinelli2019you}. We recall the important details again in this section.

We first conduct quasi-random hyperparameter search via a Sobol sequence, which aims to distribute hyperparameter settings evenly to avoid "clumping" effects \cite{bergstra2012random}. More specifically, for each dataset
and model, we generated 30 different configurations per valid combination of training type and loss function. we added a short Bayesian optimization
phase (best configuration so far + 30 new trials) to tune the hyperparameters further. All above steps are conducted using Ax framework (\url{https://ax.dev/})

We use a large hyperparameter space including loss functions (pairwise margin ranking with hinge loss, binary cross entropy, cross entropy), regularization techniques (none/L1/L2/L3, dropout), optimizers (Adam, Adagrad), and initialization methods used in the KGE community as hyperparameters. We consider 128, 256, 512 as possible embedding sizes. More details see in \citet[Table 5]{ruffinelli2019you}.

The hyperparameters of the baseline models are located within the software folder we submitted. Concretely, all configuration files (*.yaml) that we use for training baseline models/competing models/models for aggregation can be found in folder "configs".

\subsection{GPU Hours}
We use a Linux machine with a 40GB NVIDIA A100 SXM4 GPU. For each KGE methods on one benchmark dataset, we allocate at most 80 hours to fit the baseline models, 14 hours to construct competing models and 10 hours to fit the models used for aggregation. 


\section{More Experiment Results}
Due to the page limit, we represent more experiment results in this section. 
\subsection{Experiments for Link Prediction in Context of Recommendation}\label{app:full_results}
Table \ref{tab:full_main_result} presents accuracy and predictive multiplicity metrics for six KGE models across four datasets, extending the findings from Table \ref{tab:main_result}. Key observations are discussed in Section \ref{sec:measure_PM}. Notably, datasets with data leakage, such as WN18 and FB15k, consistently exhibit larger predictive multiplicity metrics compared to datasets without this issue, namely WN18RR and FB15k237. This trend is visualized in Figure \ref{fig:inference} and elaborated upon in Section \ref{sec:influence}.

\begin{table*}[h!]
\centering
\resizebox{\textwidth}{!}{%
\begin{tabular}{c|ccccc|ccccc}
\toprule[2pt]
Model & Dataset & Baselines & $Hits@10\uparrow$ & $\alpha_\epsilon\downarrow$ & $\delta_\epsilon\downarrow$ & Dataset & Baselines & $Hits@10\uparrow$ & $\alpha_\epsilon\downarrow$ & $\delta_\epsilon\downarrow$ \\\midrule
 \multirow{8}{*}{\rotatebox{90}{TransE}} & \multirow{4}{*}{\rotatebox{90}{WN18}} & w/o & 0.903 & 0.074 & 0.029 & \multirow{4}{*}{\rotatebox{90}{FB15k}} & w/o & 0.755 & 0.140 & 0.053 \\
 &  & major & 0.296 & 0.109 & 0.051 &  & major & 0.202 & 0.150 & 0.070 \\
 &  & Borda & 0.876 & 0.028 & 0.011 &  & Borda & 0.751 & 0.036 & {\ul 0.014} \\
 &  & range & {\ul 0.907} & {\ul 0.017} & {\ul 0.009} &  & range & {\ul 0.760} & {\ul 0.032} & {\ul 0.014} \\\cmidrule(lr){2-11}
 & \multirow{4}{*}{\rotatebox{90}{WN18RR}} & w/o & 0.518 & 0.076 & 0.034 & \multirow{4}{*}{\rotatebox{90}{FB15k237}} & w/o & 0.455 & 0.385 & 0.145 \\
 &  & major & 0.055 & 0.096 & 0.045 &  & major & 0.155 & 0.171 & 0.081 \\
 &  & Borda & 0.482 & 0.032 & 0.016 &  & Borda & 0.456 & 0.110 & 0.044 \\
 &  & range & {\ul 0.519} & {\ul \textbf{0.017}} & {\ul \textbf{0.009}} &  & range & {\ul 0.470} & {\ul 0.101} & {\ul 0.041} \\\midrule
 
 \multirow{8}{*}{\rotatebox{90}{RotatE}} & \multirow{4}{*}{\rotatebox{90}{WN18}} & w/o & 0.951 & 0.026 & 0.009 & \multirow{4}{*}{\rotatebox{90}{FB15k}} & w/o & 0.790 & 0.086 & 0.032 \\
 &  & major & 0.880 & 0.031 & 0.016 &  & major & 0.464 & 0.088 & 0.044 \\
 &  & Borda & {\ul \textbf{0.957}} & {\ul \textbf{0.008}} & {\ul \textbf{0.004}} &  & Borda & 0.797 & 0.018 & 0.008 \\
 &  & range & {\ul \textbf{0.957}} & {\ul \textbf{0.008}} & {\ul \textbf{0.004}} &  & range & {\ul \textbf{0.798}} & {\ul \textbf{0.016}} & {\ul \textbf{0.007}} \\\cmidrule(lr){2-11}
 & \multirow{4}{*}{\rotatebox{90}{WN18RR}} & w/o & 0.547 & 0.195 & 0.074 & \multirow{4}{*}{\rotatebox{90}{FB15k237}} & w/o & 0.520 & 0.163 & 0.064 \\
 &  & major & 0.413 & 0.064 & 0.029 &  & major & 0.204 & 0.104 & 0.053 \\
 &  & Borda & 0.564 & 0.062 & 0.028 &  & Borda & 0.523 & 0.039 & 0.017 \\
 &  & range & {\ul \textbf{0.578}} & {\ul 0.051} & {\ul 0.022} &  & range & {\ul \textbf{0.524}} & {\ul \textbf{0.037}} & {\ul \textbf{0.016}} \\\midrule
 
 \multirow{8}{*}{\rotatebox{90}{RESCAL}} & \multirow{4}{*}{\rotatebox{90}{WN18}} & w/o & 0.940 & 0.039 & 0.016 & \multirow{4}{*}{\rotatebox{90}{FB15k}} & w/o & 0.714 & 0.217 & 0.081 \\
 &  & major & 0.462 & 0.015 & 0.007 &  & major & 0.137 & 0.050 & 0.024 \\
 &  & Borda & 0.935 & {\ul 0.011} & {\ul 0.005} &  & Borda & 0.716 & 0.054 & 0.022 \\
 &  & range & {\ul 0.944} & 0.012 & {\ul 0.005} &  & range & {\ul 0.729} & {\ul 0.048} & {\ul 0.020} \\\cmidrule(lr){2-11}
 & \multirow{4}{*}{\rotatebox{90}{WN18RR}} & w/o & 0.517 & 0.248 & 0.095 & \multirow{4}{*}{\rotatebox{90}{FB15k237}} & w/o & 0.482 & 0.375 & 0.140 \\
 &  & major & 0.198 & 0.108 & 0.054 &  & major & 0.145 & 0.165 & 0.089 \\
 &  & Borda & 0.561 & 0.099 & 0.043 &  & Borda & 0.485 & 0.107 & 0.048 \\
 &  & range & {\ul 0.575} & {\ul 0.084} & {\ul 0.034} &  & range & {\ul 0.498} & {\ul 0.098} & {\ul 0.042} \\\midrule
 
\multirow{8}{*}{\rotatebox{90}{DistMult}} & \multirow{4}{*}{\rotatebox{90}{WN18}} & w/o & 0.938 & 0.044 & 0.018 & \multirow{4}{*}{\rotatebox{90}{FB15k}} & w/o & 0.773 & 0.170 & 0.064 \\
 &  & major & 0.459 & {\ul 0.012} & 0.008 &  & major & 0.157 & 0.049 & 0.023 \\
 &  & Borda & 0.927 & 0.013 & {\ul 0.006} &  & Borda & 0.766 & 0.052 & 0.021 \\
 &  & range & {\ul 0.941} & 0.015 & 0.007 &  & range & {\ul 0.778} & {\ul 0.048} & {\ul 0.019} \\\cmidrule(lr){2-11}
 & \multirow{4}{*}{\rotatebox{90}{WN18RR}} & w/o & 0.526 & 0.169 & 0.068 & \multirow{4}{*}{\rotatebox{90}{FB15k237}} & w/o & 0.476 & 0.320 & 0.120 \\
 &  & major & 0.185 & 0.078 & 0.037 &  & major & 0.144 & 0.124 & 0.059 \\
 &  & Borda & 0.524 & 0.055 & 0.024 &  & Borda & 0.475 & 0.088 & 0.037 \\
 &  & range & {\ul 0.542} & {\ul 0.048} & {\ul 0.021} &  & range & {\ul 0.488} & {\ul 0.082} & {\ul 0.034} \\\midrule
 
 \multirow{8}{*}{\rotatebox{90}{ComplEx}} & \multirow{4}{*}{\rotatebox{90}{WN18}} & w/o & 0.941 & 0.042 & 0.018 & \multirow{4}{*}{\rotatebox{90}{FB15k}} & w/o & 0.765 & 0.210 & 0.081 \\
 &  & major & 0.458 & {\ul 0.009} & {\ul 0.005} &  & major & 0.158 & {\ul 0.047} & {\ul 0.023} \\
 &  & Borda & 0.943 & 0.021 & 0.010 &  & Borda & 0.765 & 0.076 & 0.032 \\
 &  & range & {\ul 0.945} & 0.020 & 0.009 &  & range & {\ul 0.780} & 0.071 & 0.029 \\\cmidrule(lr){2-11}
 & \multirow{4}{*}{\rotatebox{90}{WN18RR}} & w/o & 0.541 & 0.217 & 0.085 & \multirow{4}{*}{\rotatebox{90}{FB15k237}} & w/o & 0.482 & 0.308 & 0.116 \\
 &  & major & 0.243 & 0.243 & 0.126 &  & major & 0.145 & 0.121 & 0.055 \\
 &  & Borda & 0.559 & 0.067 & 0.030 &  & Borda & 0.480 & 0.087 & 0.036 \\
 &  & range & {\ul 0.573} & {\ul 0.058} & {\ul 0.024} &  & range & {\ul 0.493} & {\ul 0.082} & {\ul 0.032} \\\midrule
 
\multirow{8}{*}{\rotatebox{90}{ConvE}} & \multirow{4}{*}{\rotatebox{90}{WN18}} & w/o & 0.938 & 0.043 & 0.019 & \multirow{4}{*}{\rotatebox{90}{FB15k}} & w/o & 0.766 & 0.177 & 0.066 \\
 &  & major & 0.476 & 0.039 & 0.021 &  & major & 0.175 & 0.085 & 0.041 \\
 &  & Borda & 0.933 & {\ul 0.014} & {\ul 0.005} &  & Borda & 0.761 & 0.052 & 0.022 \\
 &  & range & {\ul 0.942} & 0.015 & 0.006 &  & range & {\ul 0.771} & {\ul 0.049} & {\ul 0.020} \\\cmidrule(lr){2-11}
 & \multirow{4}{*}{\rotatebox{90}{WN18RR}} & w/o & 0.500 & 0.222 & 0.088 & \multirow{4}{*}{\rotatebox{90}{FB15k237}} & w/o & 0.474 & 0.340 & 0.130 \\
 &  & major & 0.185 & 0.092 & 0.047 &  & major & 0.150 & 0.154 & 0.074 \\
 &  & Borda & 0.522 & 0.082 & 0.035 &  & Borda & 0.474 & 0.092 & 0.039 \\
 &  & range & {\ul 0.534} & {\ul 0.068} & {\ul 0.027} &  & range & {\ul 0.486} & {\ul 0.085} & {\ul 0.034}\\ \bottomrule[2pt]
\end{tabular}
}
\caption{This table presents the metrics for accuracy (i.e. Hits@K and $\epsilon$) and for predictive multiplicity (i.e. $\alpha_\epsilon$ and $\delta_\epsilon$) for different voting methods applied on different KGE models and four datasets. }\label{tab:full_main_result}
\end{table*}

\subsection{Experiments for Link Prediction in Context of Query Answering}\label{app:query_answering}
We define link prediction as binary classification problem in the main body of the paper, it is suitable for recommendation systems, where people only care about the top-K results. But there are cases where people care more about the answer set of the query. For example, CQD \cite{Arakelyan2021cqd} decomposite logical queries into one-step atomic queries like $\langle h,r,?\rangle$ or $\langle ?,r,t\rangle$ and predict the answer set for each atomic query with ComplEx. In this case, We can define link prediction as predicting an answer set $A$ for queries. We denote $tr(q,e)$ as the corresponding triple $\langle h,r,e\rangle$ or $\langle e,r,t\rangle$, respectively. 
\begin{definition}[Link Prediction for Query Answering]
Given a KGE model $M_\theta$, a query $q\in Q$ and a scoring-based threshold $\tau$, the answer set $A$ of the query $q$ include all entities that have predictive scores exceeding the threshold.
    \begin{equation}
        A_{\tau}(M_\theta, q)=\{e\in E \mid M_\theta(tr(q,e))\geq\tau\}.
    \end{equation}
\end{definition}

Then we adapt all definition of predictive multiplicity and its metrics to this setting. The definition of the $\epsilon$-level set remains the same. Embedding-based query answering exhibits predictive multiplicity if competing models suggest different answer sets for a given query. 
\begin{definition}[Predictive Multiplicity]\label{def:pm_qe}
    Given a threshold $\tau$, a baseline model $M_{\theta}^*$, and an error tolerance $\epsilon$, the prediction of query $q$ exhibits predictive multiplicity if there exists a model $M_\theta\in S_\epsilon(M_{\theta}^*)$ such that $A_{\tau}(M_{\theta}, q)\not = A_{\tau}(M_{\theta}^*, q)$.
\end{definition}

\begin{table}[h!]
\centering
\resizebox{0.45\textwidth}{!}{%
\begin{tabular}{c|ccccc|ccccc}
\toprule[2pt]
model & dataset & baseline & Hits@1$\uparrow$ & $\alpha$@1$\downarrow$ & $\delta$@1$\downarrow$ & dataset & baseline & Hits@1$\uparrow$ & $\alpha$@1$\downarrow$ & $\delta$@1$\downarrow$ \\\midrule
\multirow{8}{*}{\rotatebox{90}{TransE}} & \multirow{4}{*}{\rotatebox{90}{WN18}} & w/o & {\ul 0.499} & 0.459 & 0.315 & \multirow{4}{*}{\rotatebox{90}{FB15k}} & w/o & 0.659 & 0.376 & 0.273 \\
 &  & major & 0.494 & 0.210 & 0.145 &  & major & 0.654 & 0.180 & 0.118 \\
 &  & Borda & 0.494 & 0.203 & 0.135 &  & Borda & 0.655 & 0.159 & 0.115 \\
 &  & range & 0.497 & {\ul \textbf{0.194}} & {\ul \textbf{0.129}} &  & range & {\ul 0.661} & {\ul \textbf{0.144}} & {\ul \textbf{0.100}} \\\cmidrule(lr){2-11}
 & \multirow{4}{*}{\rotatebox{90}{WN18RR}} & w/o & 0.105 & 0.647 & 0.472 & \multirow{4}{*}{\rotatebox{90}{FB15k237}} & w/o & 0.544 & 0.312 & 0.206 \\
 &  & major & 0.106 & 0.318 & 0.218 &  & major & 0.543 & 0.101 & 0.069 \\
 &  & Borda & 0.109 & 0.308 & 0.228 &  & Borda & 0.544 & {\ul 0.090} & {\ul 0.060} \\
 &  & range & {\ul 0.112} & {\ul 0.283} & {\ul 0.215} &  & range & {\ul 0.548} & 0.091 & 0.061 \\\midrule
 
\multirow{8}{*}{\rotatebox{90}{RotatE}} & \multirow{4}{*}{\rotatebox{90}{WN18}} & w/o & {\ul \textbf{0.880}} & 0.413 & 0.343 & \multirow{4}{*}{\rotatebox{90}{FB15k}} & w/o & 0.664 & 0.473 & 0.376 \\
 &  & major & 0.872 & 0.300 & 0.219 &  & major & 0.674 & 0.271 & 0.198 \\
 &  & Borda & 0.873 & 0.296 & 0.229 &  & Borda & 0.681 & 0.263 & 0.215 \\
 &  & range & 0.875 & {\ul 0.285} & {\ul 0.218} &  & range & {\ul 0.682} & {\ul 0.248} & {\ul 0.188} \\\cmidrule(lr){2-11}
 & \multirow{4}{*}{\rotatebox{90}{WN18RR}} & w/o & 0.219 & 0.415 & 0.297 & \multirow{4}{*}{\rotatebox{90}{FB15k237}} & w/o & 0.539 & 0.335 & 0.221 \\
 &  & major & 0.226 & 0.172 & 0.115 &  & major & 0.538 & 0.095 & 0.063 \\
 &  & Borda & 0.224 & 0.187 & 0.130 &  & Borda & 0.540 & {\ul \textbf{0.089}} & {\ul \textbf{0.057}} \\
 &  & range & {\ul 0.231} & {\ul \textbf{0.145}} & {\ul \textbf{0.100}} &  & range & {\ul 0.543} & 0.096 & 0.066 \\\midrule
 
\multirow{8}{*}{\rotatebox{90}{RESCAL}} & \multirow{4}{*}{\rotatebox{90}{WN18}} & w/o & 0.785 & 0.647 & 0.483 & \multirow{4}{*}{\rotatebox{90}{FB15k}} & w/o & 0.537 & 0.605 & 0.496 \\
 &  & major & 0.862 & 0.344 & {\ul 0.250} &  & major & 0.584 & 0.402 & {\ul 0.282} \\
 &  & Borda & 0.867 & 0.340 & 0.274 &  & Borda & 0.595 & 0.385 & 0.309 \\
 &  & range & {\ul 0.868} & {\ul 0.331} & 0.263 &  & range & {\ul 0.605} & {\ul 0.374} & 0.301 \\\cmidrule(lr){2-11}
 & \multirow{4}{*}{\rotatebox{90}{WN18RR}} & w/o & 0.194 & 0.734 & 0.639 & \multirow{4}{*}{\rotatebox{90}{FB15k237}} & w/o & 0.492 & 0.635 & 0.518 \\
 &  & major & 0.213 & 0.510 & 0.372 &  & major & 0.534 & 0.352 & 0.249 \\
 &  & Borda & 0.232 & 0.425 & 0.332 &  & Borda & 0.550 & 0.326 & 0.252 \\
 &  & range & {\ul \textbf{0.241}} & {\ul 0.395} & {\ul 0.318} &  & range & {\ul 0.556} & {\ul 0.308} & {\ul 0.231} \\\midrule
 
\multirow{8}{*}{\rotatebox{90}{DistMult}} & \multirow{4}{*}{\rotatebox{90}{WN18}} & w/o & 0.861 & 0.385 & 0.325 & \multirow{4}{*}{\rotatebox{90}{FB15k}} & w/o & 0.696 & 0.425 & 0.349 \\
 &  & major & 0.860 & {\ul 0.298} & {\ul 0.208} &  & major & 0.695 & 0.306 & {\ul 0.219} \\
 &  & Borda & {\ul 0.862} & 0.310 & 0.247 &  & Borda & 0.697 & 0.302 & 0.247 \\
 &  & range & {\ul 0.862} & 0.304 & 0.238 &  & range & {\ul \textbf{0.700}} & {\ul 0.298} & 0.240 \\\cmidrule(lr){2-11}
 & \multirow{4}{*}{\rotatebox{90}{WN18RR}} & w/o & 0.133 & 0.780 & 0.675 & \multirow{4}{*}{\rotatebox{90}{FB15k237}} & w/o & 0.417 & 0.817 & 0.643 \\
 &  & major & 0.163 & 0.517 & 0.366 &  & major & 0.511 & 0.392 & 0.271 \\
 &  & Borda & 0.171 & 0.449 & 0.342 &  & Borda & 0.523 & 0.399 & 0.297 \\
 &  & range & {\ul 0.183} & {\ul 0.412} & {\ul 0.293} &  & range & {\ul 0.543} & {\ul 0.330} & {\ul 0.241} \\\midrule
 
\multirow{8}{*}{\rotatebox{90}{ComplEx}} & \multirow{4}{*}{\rotatebox{90}{WN18}} & w/o & 0.866 & 0.379 & 0.325 & \multirow{4}{*}{\rotatebox{90}{FB15k}} & w/o & 0.685 & 0.420 & 0.350 \\
 &  & major & 0.866 & 0.310 & {\ul 0.222} &  & major & 0.694 & {\ul 0.272} & {\ul 0.206} \\
 &  & Borda & 0.866 & {\ul 0.305} & 0.238 &  & Borda & 0.698 & 0.274 & 0.218 \\
 &  & range & {\ul 0.867} & 0.308 & 0.238 &  & range & {\ul \textbf{0.700}} & 0.273 & 0.217 \\\cmidrule(lr){2-11}
 & \multirow{4}{*}{\rotatebox{90}{WN18RR}} & w/o & 0.100 & 0.957 & 0.876 & \multirow{4}{*}{\rotatebox{90}{FB15k237}} & w/o & 0.419 & 0.820 & 0.650 \\
 &  & major & 0.158 & 0.702 & 0.487 &  & major & 0.509 & 0.393 & 0.276 \\
 &  & Borda & 0.194 & 0.553 & 0.438 &  & Borda & 0.528 & 0.398 & 0.297 \\
 &  & range & {\ul 0.203} & {\ul 0.489} & {\ul 0.372} &  & range & {\ul 0.543} & {\ul 0.332} & {\ul 0.256} \\\midrule
 
\multirow{8}{*}{\rotatebox{90}{ConvE}} & \multirow{4}{*}{\rotatebox{90}{WN18}} & w/o & {\ul 0.870} & 0.435 & 0.354 & \multirow{4}{*}{\rotatebox{90}{FB15k}} & w/o & 0.634 & 0.568 & 0.436 \\
 &  & major & 0.862 & 0.302 & {\ul 0.214} &  & major & 0.672 & 0.315 & {\ul 0.219} \\
 &  & Borda & 0.863 & 0.309 & 0.251 &  & Borda & 0.686 & 0.307 & 0.240 \\
 &  & range & 0.864 & {\ul 0.299} & 0.239 &  & range & {\ul 0.688} & {\ul 0.293} & 0.225 \\\cmidrule(lr){2-11}
 & \multirow{4}{*}{\rotatebox{90}{WN18RR}} & w/o & 0.150 & 0.617 & 0.469 & \multirow{4}{*}{\rotatebox{90}{FB15k237}} & w/o & 0.520 & 0.554 & 0.434 \\
 &  & major & 0.164 & 0.332 & 0.232 &  & major & 0.538 & 0.283 & 0.194 \\
 &  & Borda & 0.164 & 0.324 & 0.240 &  & Borda & 0.548 & 0.255 & 0.178 \\
 &  & range & {\ul 0.168} & {\ul 0.288} & {\ul 0.201} &  & range & {\ul 0.553} & {\ul 0.218} & {\ul 0.145}\\\bottomrule[2pt]
\end{tabular}
}
\caption{predictive multiplicity evaluation for top-1 answers in query answering setting. }\label{tab:query_asnwering_top1}
\end{table}

\begin{definition}[Ambiguity]
    Given a testing query set $Q'$ and a threshold $\tau$, the ambiguity of link prediction over the $\epsilon$-level set $S_\epsilon(M_{\theta}^*)$ is the proportion of testing queries that are  provided different answer sets by a competing model $M_\theta\in S_\epsilon(M_{\theta}^*)$:
    \begin{equation}
        \small\alpha(M_{\theta}^*):=\frac{1}{|Q'|}\sum_{q\in Q'}\max_{M_\theta\in\mathcal{M}}\mathbbm{1}[A_\tau(M_\theta, q)\not = A_\tau(M_\theta^*, q)].
    \end{equation}
\end{definition}

\begin{definition}[Discrepancy]
    Given a testing query set $Q'$ and a threshold $\tau$, the discrepancy of link prediction over the $\epsilon$-level set $S_\epsilon(M_{\theta}^*)$ is the maximum proportion of testing queries that are provided different answer sets by a competing model $M_\theta\in S_\epsilon(M_{\theta}^*)$:
    \begin{equation}
        \small\delta(M_{\theta}^*):=\max_{M_\theta\in\mathcal{M}}\frac{1}{|Q'|}\sum_{q\in Q'}\mathbbm{1}[A_\tau(M_\theta, q)\not = A_\tau(M_\theta^*, q)].
    \end{equation}
\end{definition}

Additionally, we introduce a new evaluation metric \textit{agreement} to measure the overlap of the predicted answer sets from competing models based on Jaccard similarity \cite{jaccard1901distribution}. The Jaccard similarity \cite{jaccard1901distribution} between two sets, denoted as $Sim(A,B)$, is defined as the ratio of the cardinality of their intersection to the cardinality of their union.
\begin{equation}
    Sim(A,B):= \frac{|A\cap B|}{|A\cup B|}
\end{equation}

Agreement is then defined as
\begin{definition}[Agreement]
    Given a testing query set $Q'$ and a threshold $\tau$, the agreement of link prediction over the $\epsilon$-level set $S_\epsilon(M_{\theta}^*)$ is average Jaccard similarity of predicted answer sets provided by competing models $M_\theta\in S_\epsilon(M_{\theta}^*)$.
    \begin{align*}
        &J(M_{\theta}^*)= \\
        &\frac{\sum_{q\in Q'}\sum_{M_\theta\in S_\epsilon(M_{\theta}^*)}Sim(P_\tau(M_\theta, q),P_\tau(M_\theta^*, q))}{|Q'|\cdot|S_\epsilon(M_{\theta}^*)|}
    \end{align*}
\end{definition}

We summarize the results of multiplicity in this setting in Table \ref{tab:query_asnwering_top1} and \ref{tab:query_asnwering_top10}. We observe more significant predictive multiplicity behavior, since it is more challenging to predict the same answer set from competing models. It requires very robust rankings from competing models. And it heavily relies on the scoring-based threshold. Nevertheless, voting method reduce the number of conflicting prediction also in that settings. In the future work, it is interesting to find out a way to set the threshold properly or at least quantify the uncertainty of the answer set for the threshold. 

\begin{table*}[h!]
\centering
\resizebox{0.95\textwidth}{!}{%
\begin{tabular}{c|cccccc|cccccc}
\toprule[2pt]
model & dataset & baseline & Hits@10$\uparrow$ & $\alpha$@10$\downarrow$ & $\delta$@10$\downarrow$ & J@10$\uparrow$ & dataset & baseline & Hits@10$\uparrow$ & $\alpha$@10$\downarrow$ & $\delta$@10$\downarrow$ & J@10$\uparrow$ \\\midrule
\multirow{8}{*}{\rotatebox{90}{TransE}} & \multirow{4}{*}{\rotatebox{90}{WN18}} & w/o & {\ul 0.662} & 0.940 & 0.825 & 0.727 & \multirow{4}{*}{\rotatebox{90}{FB15k}} & w/o & 0.468 & 0.959 & 0.891 & 0.597\\
 &  & major & 0.088 & {\ul 0.480} & {\ul 0.376} & {\ul 0.916} & & major & 0.133 & {\ul 0.545} & {\ul 0.469} & {\ul 0.878}\\
 &  & Borda & 0.522 & 0.673 & 0.505 & 0.871 & & Borda & 0.463 & 0.702 & 0.570 & 0.850\\
 &  & range & 0.522 & 0.649 & 0.497 & 0.876 & & range & {\ul 0.464} & 0.683 & 0.549 & 0.859\\\cmidrule(lr){2-13}
 & \multirow{4}{*}{\rotatebox{90}{WN18RR}} & w/o & 0.517 & 0.990 & 0.930 & 0.650 & \multirow{4}{*}{\rotatebox{90}{FB15k237}} & w/o & 0.239 & 0.991 & 0.952 & 0.564\\
 &  & major & 0.106 & {\ul 0.226} & {\ul 0.177} & {\ul 0.961} & & major & 0.072 & {\ul 0.536} & {\ul 0.447} & {\ul 0.906} \\
 &  & Borda & 0.659 & 0.560 & 0.428 & 0.910 & & Borda & {\ul0.242} & 0.770 & 0.633 & 0.865\\
 &  & range & {\ul 0.660} & 0.529 & 0.401 & 0.916 & & range & {\ul0.242} & 0.751 & 0.611 & 0.873\\\midrule
 
\multirow{8}{*}{\rotatebox{90}{RotatE}} & \multirow{4}{*}{\rotatebox{90}{WN18}} & w/o & {\ul 0.730} & 0.986 & 0.978 & 0.391 & \multirow{4}{*}{\rotatebox{90}{FB15k}} & w/o & {\ul 0.435} & 0.967 & 0.934 & 0.441 \\
 &  & major & 0.441 & {\ul 0.355} & {\ul 0.298} & {\ul 0.917} & & major & 0.114 & {\ul 0.654} & {\ul 0.590} & {\ul 0.844} \\
 &  & Borda & 0.717 & 0.902 & 0.855 & 0.719 & & Borda & {\ul 0.435} & 0.802 & 0.712 & 0.785\\
 &  & range & 0.717 & 0.868 & 0.783 & 0.747 & & range & {\ul 0.435} & 0.790 & 0.693 & 0.796\\\cmidrule(lr){2-13}
 & \multirow{4}{*}{\rotatebox{90}{WN18RR}} & w/o & 0.540 & 0.976 & 0.935 & 0.589 & \multirow{4}{*}{\rotatebox{90}{FB15k237}} & w/o & {\ul 0.244} & 0.955 & 0.870 & 0.695 \\
 &  & major & 0.188 & {\ul 0.270} & {\ul 0.209} & {\ul 0.955} & & major & 0.058 & {\ul 0.424} & {\ul 0.331} & {\ul 0.944} \\
 &  & Borda & {\ul 0.541} & 0.765 & 0.625 & 0.856 & & Borda & {\ul 0.244} & 0.610 & 0.457 & 0.916 \\
 &  & range & {\ul 0.541} & 0.723 & 0.589 & 0.877 & & range & {\ul 0.244} & 0.586 & 0.431 & 0.922 \\\midrule

\multirow{8}{*}{\rotatebox{90}{RESCAL}} & \multirow{4}{*}{\rotatebox{90}{WN18}} & w/o & 0.671 & 1.000 & 1.000 & 0.156 & \multirow{4}{*}{\rotatebox{90}{FB15k}} & w/o & 0.345 & 0.999 & 0.989 & 0.334\\
 &  & major & 0.515 & {\ul 0.731} & {\ul 0.623} & {\ul 0.848} & & major & 0.104 & {\ul 0.776} & {\ul 0.714} & {\ul 0.753}\\
 &  & Borda & {\ul 0.713} & 0.978 & 0.951 & 0.604 & & Borda & 0.386 & 0.913 & 0.842 & 0.671\\
 &  & range & 0.708 & 0.964 & 0.918 & 0.647 & & range & {\ul 0.389} & 0.905 & 0.828 & 0.679\\\cmidrule(lr){2-13}
 & \multirow{4}{*}{\rotatebox{90}{WN18RR}} & w/o & 0.529 & 0.996 & 0.995 & 0.247 & \multirow{4}{*}{\rotatebox{90}{FB15k237}} & w/o & 0.210 & 1.000 & 0.999 & 0.236 \\
 &  & major & 0.165 & {\ul 0.440} & {\ul 0.400} & {\ul 0.862} & & major & 0.128 & 0.909 & {\ul 0.873} & {\ul 0.693} \\
 &  & Borda & 0.549 & 0.926 & 0.852 & 0.689 & & Borda & 0.239 & 0.955 & 0.897 & 0.672 \\
 &  & range & {\ul 0.550} & 0.898 & 0.817 & 0.722 & & range & {\ul 0.241} & {\ul 0.947} & 0.881 & 0.689 \\\midrule

\multirow{8}{*}{\rotatebox{90}{DistMult}} & \multirow{4}{*}{\rotatebox{90}{WN18}} & w/o & 0.701 & 0.982 & 0.972 & 0.343 & \multirow{4}{*}{\rotatebox{90}{FB15k}} & w/o & {\ul 0.439} & 0.970 & 0.936 & 0.459\\
 &  & major & 0.209 & {\ul 0.271} & {\ul 0.235} & {\ul 0.929} & & major & 0.103 & {\ul 0.697} & {\ul 0.633} & {\ul 0.784}\\
 &  & Borda & {\ul 0.702} & 0.922 & 0.872 & 0.694 & & Borda & 0.427 & 0.820 & 0.725 & 0.717\\
 &  & range & {\ul 0.702} & 0.883 & 0.802 & 0.739 & & range & 0.428 & 0.805 & 0.703 & 0.727\\\cmidrule(lr){2-13}
 & \multirow{4}{*}{\rotatebox{90}{WN18RR}} & w/o & 0.512 & 1.000 & 1.000 & 0.181 & \multirow{4}{*}{\rotatebox{90}{FB15k237}} & w/o & 0.198 & 1.000 & 1.000 & 0.183 \\
 &  & major & 0.404 & {\ul 0.534} & {\ul 0.467} & {\ul 0.850} & & major & 0.159 & {\ul 0.974} & 0.942 & {\ul 0.667} \\
 &  & Borda & 0.541 & 0.973 & 0.934 & 0.659 & & Borda & 0.243 & 0.987 & 0.958 & 0.593 \\
 &  & range & {\ul 0.543} & 0.958 & 0.892 & 0.699 & & range & {\ul 0.248} & {\ul 0.974} & {\ul 0.935} & 0.634 \\\midrule

\multirow{8}{*}{\rotatebox{90}{ComplEx}} & \multirow{4}{*}{\rotatebox{90}{WN18}} & w/o & {\ul 0.716} & 0.985 & 0.973 & 0.409 & \multirow{4}{*}{\rotatebox{90}{FB15k}} & w/o & 0.420 & 0.957 & 0.925 & 0.427\\
 &  & major & 0.196 & {\ul 0.248} & {\ul 0.220} & {\ul 0.928} & & major & 0.061 & {\ul 0.635} & {\ul 0.582} & {\ul 0.801}\\
 &  & Borda & 0.705 & 0.876 & 0.784 & 0.760 & & Borda & 0.423 & 0.841 & 0.768 & 0.711\\
 &  & range & 0.705 & 0.830 & 0.731 & 0.785 & & range & {\ul 0.425} & 0.832 & 0.753 & 0.724\\\cmidrule(lr){2-13}
 & \multirow{4}{*}{\rotatebox{90}{WN18RR}} & w/o & 0.456 & 1.000 & 1.000 & 0.103 & \multirow{4}{*}{\rotatebox{90}{FB15k237}} & w/o & 0.197 & 1.000 & 1.000 & 0.172 \\
 &  & major & 0.437 & {\ul 0.911} & {\ul 0.839} & {\ul 0.720} & & major & 0.163 & {\ul 0.972} & {\ul 0.946} & {\ul 0.660} \\
 &  & Borda & 0.545 & 0.990 & 0.966 & 0.587 & & Borda & 0.246 & 0.990 & 0.961 & 0.588 \\
 &  & range & {\ul 0.549} & 0.979 & 0.941 & 0.628 & & range & {\ul 0.250} & 0.979 & 0.950 & 0.626 \\\midrule
 
\multirow{8}{*}{\rotatebox{90}{ConvE}} & \multirow{4}{*}{\rotatebox{90}{WN18}} & w/o & {\ul 0.713} & 0.993 & 0.989 & 0.277 & \multirow{4}{*}{\rotatebox{90}{FB15k}} & w/o & 0.429 & 0.998 & 0.990 & 0.354\\
 &  & major & 0.392 & {\ul 0.379} & {\ul 0.314} & {\ul 0.915} & & major & 0.196 & {\ul 0.800} & {\ul 0.721} & {\ul 0.780}\\
 &  & Borda & 0.704 & 0.939 & 0.877 & 0.708 & & Borda & 0.439 & 0.910 & 0.833 & 0.709\\
 &  & range & 0.705 & 0.912 & 0.825 & 0.745 & & range & {\ul 0.440} & 0.892 & 0.806 & 0.732\\\cmidrule(lr){2-13}
 & \multirow{4}{*}{\rotatebox{90}{WN18RR}} & w/o & 0.527 & 0.995 & 0.989 & 0.351 & \multirow{4}{*}{\rotatebox{90}{FB15k237}} & w/o & 0.236 & 0.999 & 0.989 & 0.370 \\
 &  & major & 0.152 & {\ul 0.393} & {\ul 0.335} & {\ul 0.913} & & major & 0.108 & {\ul 0.815} & {\ul 0.747} & {\ul 0.784} \\
 &  & Borda & {\ul 0.537} & 0.905 & 0.813 & 0.739 & & Borda & 0.249 & 0.909 & 0.813 & 0.761 \\
 &  & range & 0.535 & 0.873 & 0.770 & 0.775 & & range & {\ul 0.251} & 0.893 & 0.788 & 0.781 \\\midrule
\end{tabular}
}
\caption{predictive multiplicity evaluation for top-10 answers in query answering setting. }\label{tab:query_asnwering_top10}
\end{table*}

\subsection{Accuracy for ComplEx on Nations dataset with respect to $\epsilon$}\label{app:acc}
\begin{figure}[h!]
\centering
\includegraphics[width=.48\textwidth]{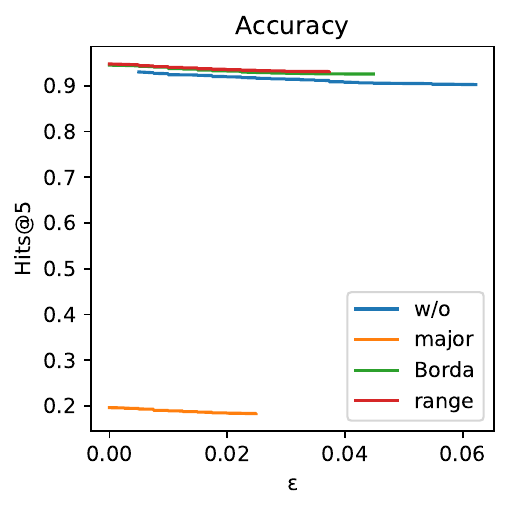}
\caption{Accuracy for ComplEx on Nations dataset with respect to $\epsilon$.}\label{fig:epsilon_curve_acc}
\end{figure}

See figure \ref{fig:epsilon_curve_acc}.

\subsection{Complete Results of Investigating the Number of Aggregated Models}\label{app:agg_exp}
\begin{figure}[h!]
\centering
\includegraphics[width=.48\textwidth]{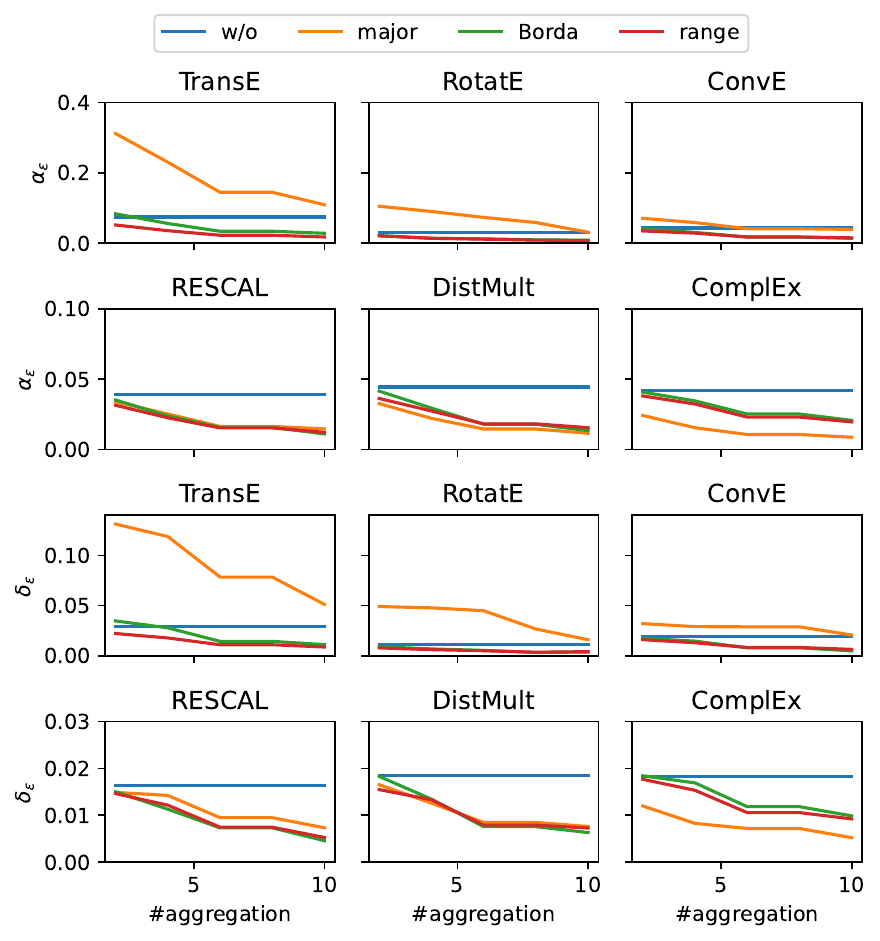}
\caption{Investigation on WN18. }\label{fig:agg_exp_wn18}
\end{figure}

\begin{figure}[h!]
\centering
\includegraphics[width=.48\textwidth]{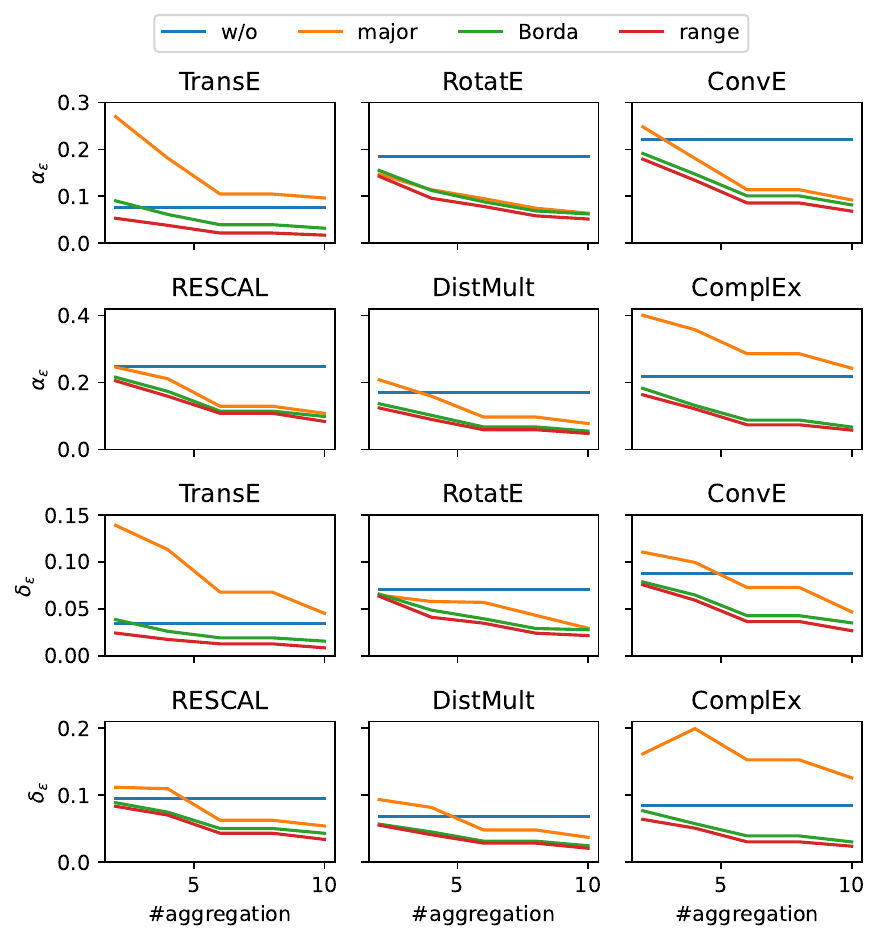}
\caption{Investigation on WN18RR. }\label{fig:agg_exp_wn18rr}
\end{figure}

\begin{figure}[h!]
\centering
\includegraphics[width=.48\textwidth]{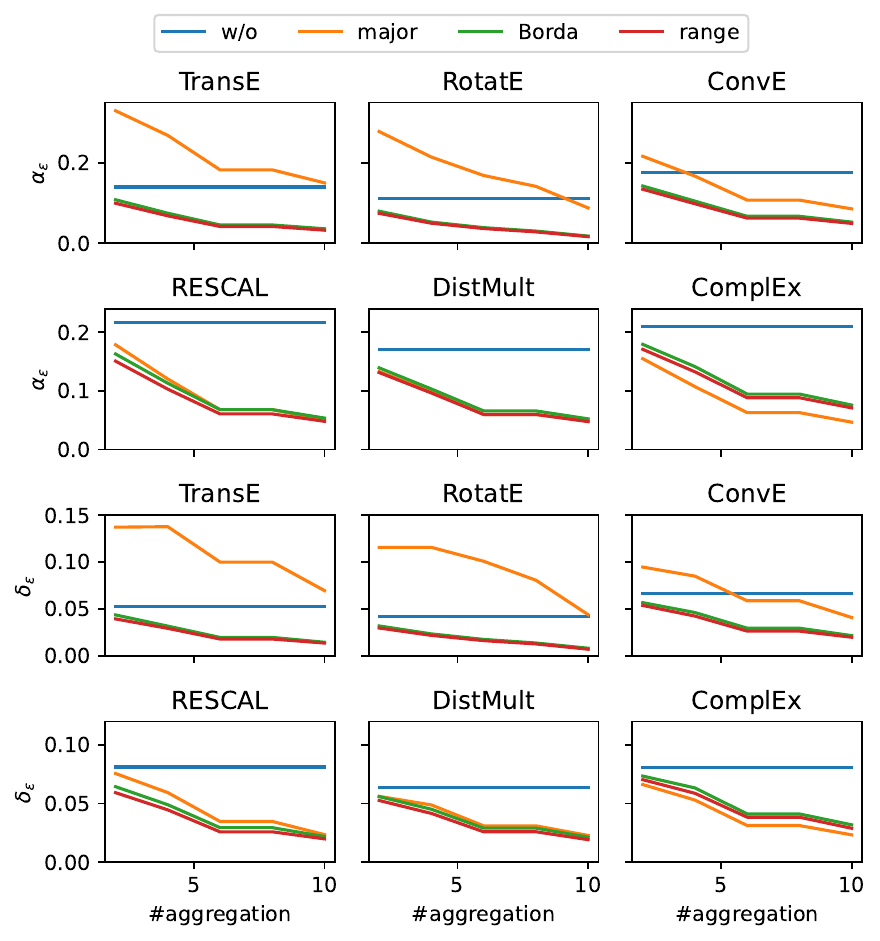}
\caption{Investigation on FB15k. }\label{fig:agg_exp_fb15k}
\end{figure}

\begin{figure}[h!]
\centering
\includegraphics[width=.48\textwidth]{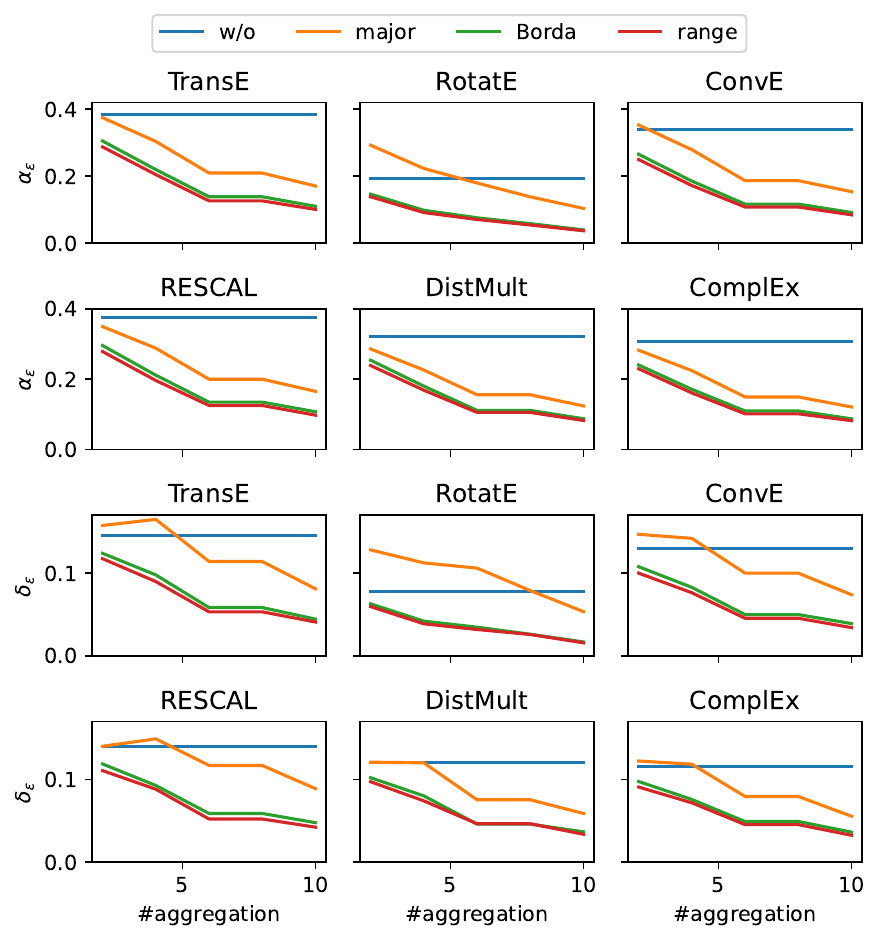}
\caption{Investigation on FB15k237. }\label{fig:agg_exp_fb15k237}
\end{figure}

\begin{figure}[h!]
\centering
\includegraphics[width=.48\textwidth]{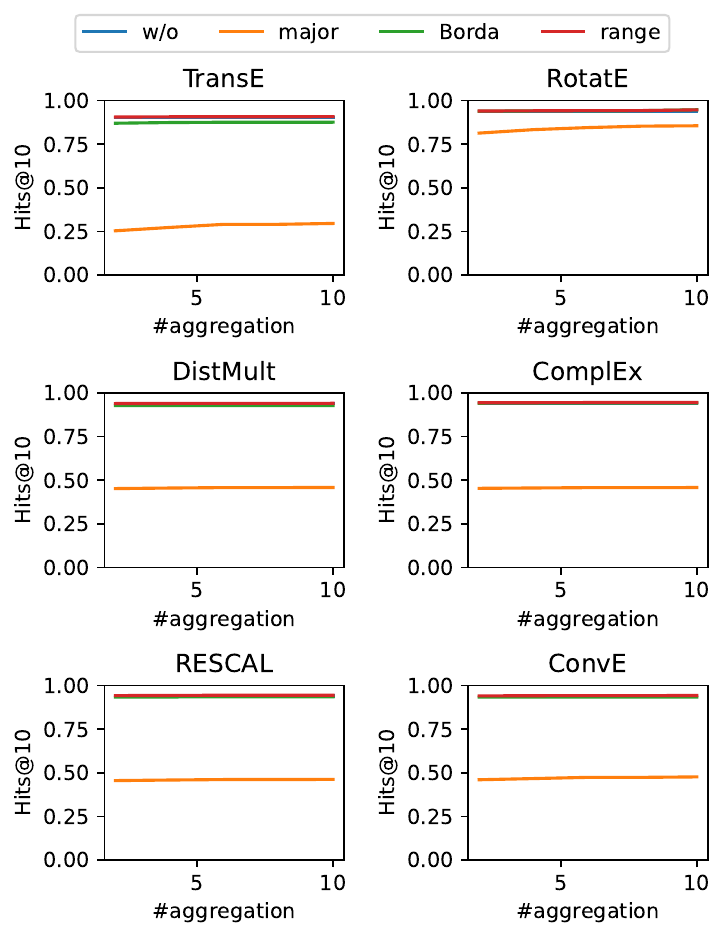}
\caption{Accuracy investigation on WN18. Note the blue lines (w/o) might be covered by other lines and not visible in diagram.}\label{fig:agg_exp_acc_wn18}
\end{figure}

\begin{figure}[h!]
\centering
\includegraphics[width=.48\textwidth]{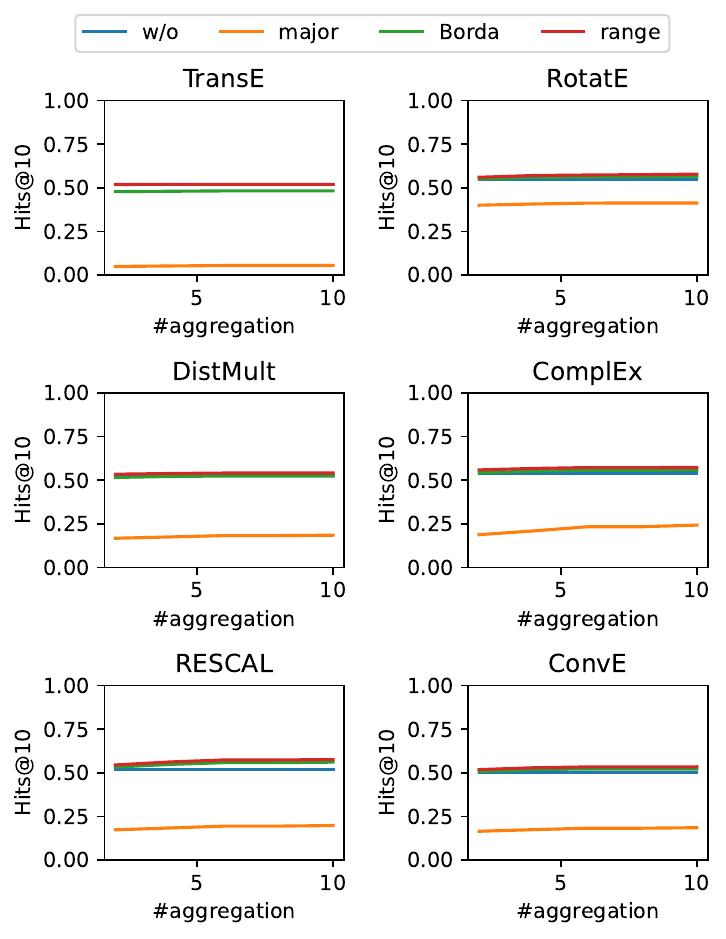}
\caption{Accuracy investigation on WN18RR. Note the blue lines (w/o) might be covered by other lines and not visible in diagram.}\label{fig:agg_exp_acc_wn18rr}
\end{figure}

\begin{figure}[h!]
\centering
\includegraphics[width=.48\textwidth]{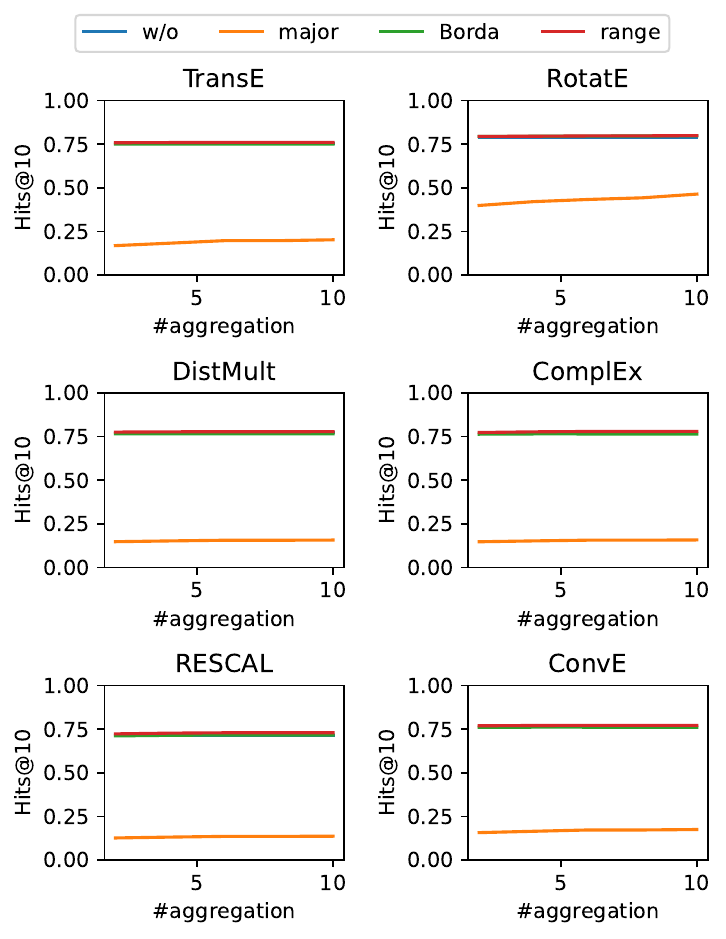}
\caption{Accuracy investigation on FB15k. Note the blue lines (w/o) might be covered by other lines and not visible in diagram.}\label{fig:agg_exp_acc_fb15k}
\end{figure}

\begin{figure}[h!]
\centering
\includegraphics[width=.48\textwidth]{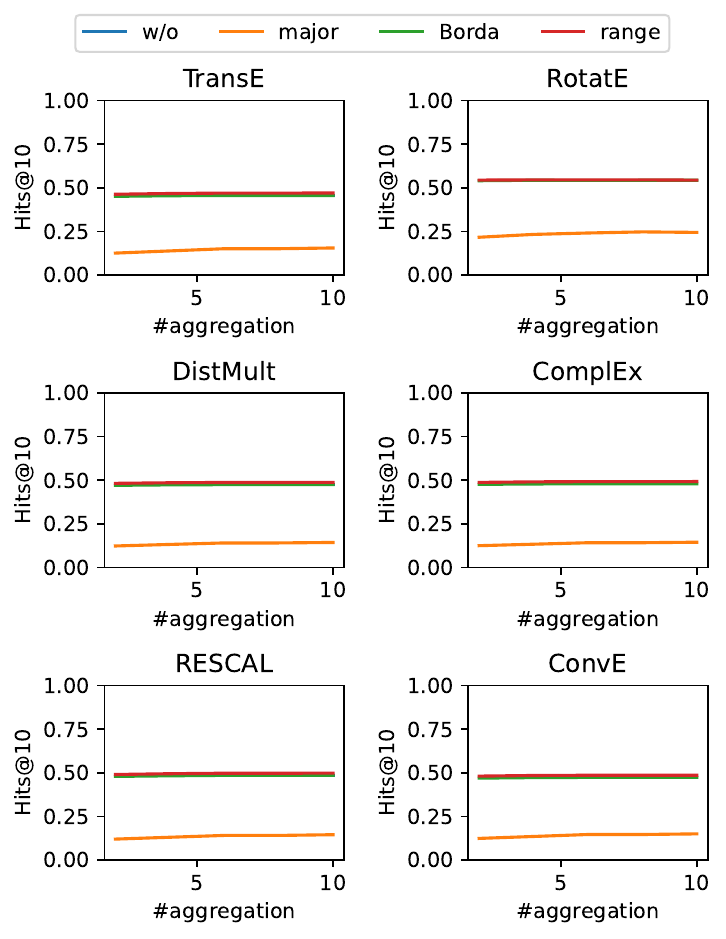}
\caption{Accuracy investigation on FB15k237. Note the blue lines (w/o) might be covered by other lines and not visible in diagram.}\label{fig:agg_exp_acc_fb15k237}
\end{figure}

Figure \ref{fig:agg_exp_wn18} - \ref{fig:agg_exp_fb15k237} show the results of investigating the predictive multiplicity wrt. the number of aggregated models for all models across all datasets. Figure \ref{fig:agg_exp_acc_wn18} - \ref{fig:agg_exp_acc_fb15k237} show the results of investigating the accuracy wrt. the number of aggregated models for all models across all datasets. 

\section{Further Discussion}
Model or predictive multiplicity is not always problematic in some Knowledge Graph Embedding (KGE) applications, as it allows flexibility in model selection. For instance, KGE methods have been used to approximate statistical reasoning on statistical $\mathcal{EL}$ ($\mathcal{SEL}$) ontologies \cite{yuqicheng2023towards, zhu2024approximating, DBLP:conf/semweb/XiongPTNS22}. In these cases, the probability intervals of axioms are estimated by assessing the point estimates from multiple KGE models trained with different random seeds. However, if the KGE models are insufficiently diverse due to random sampling from the hypothesis space, these probability intervals may be too narrow. Introducing predictive multiplicity metrics, such as ambiguity and discrepancy, to measure model diversity could improve the sampling process, potentially leading to better approximations for statistical reasoning.

Additionally, previous work has explored logical explanations for KGE predictions. For example, \citet{qu2019probabilistic, cheng2022rlogic} provide logic rules as explanations, while \citet{he2023can, he2024ecai} propose a query embedding model that explains knowledge in the form of $\mathcal{SROI}^{-}$ description logic axioms. However, due to model multiplicity, these explanations can vary depending on the random seeds used during training. Addressing this variability by finding a set of conflict-free explanations, potentially through the use of computational argumentation frameworks \cite{nico2021aaai, nico2018kr}, is another promising direction for future research. Future work can also be extend to handle knowledge graphs with high-order relational structure \cite{NestedE2024,bo2023shrinking, jingcheng1, jingcheng2}.

\subsection{Relationship between Predictive Multiplicity and Entity/Relation Frequency}\label{app:frequency}
\begin{figure}[h!]
\centering
\includegraphics[width=.48\textwidth]{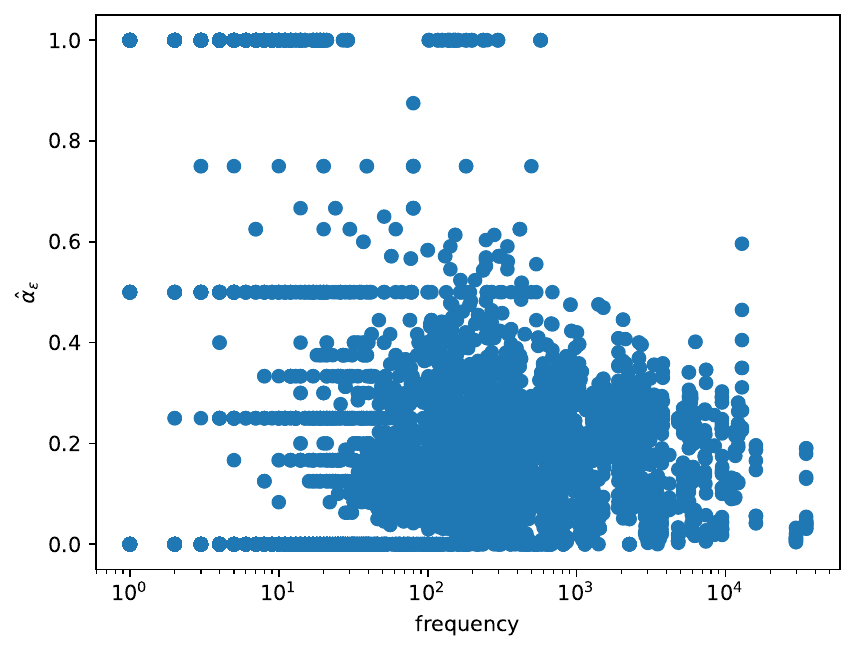}
\caption{This figure demonstrates the weak negative correlation between relation frequency and empirical ambiguity. }\label{fig:relation_alpha}
\end{figure}

\begin{figure}[h!]
\centering
\includegraphics[width=.48\textwidth]{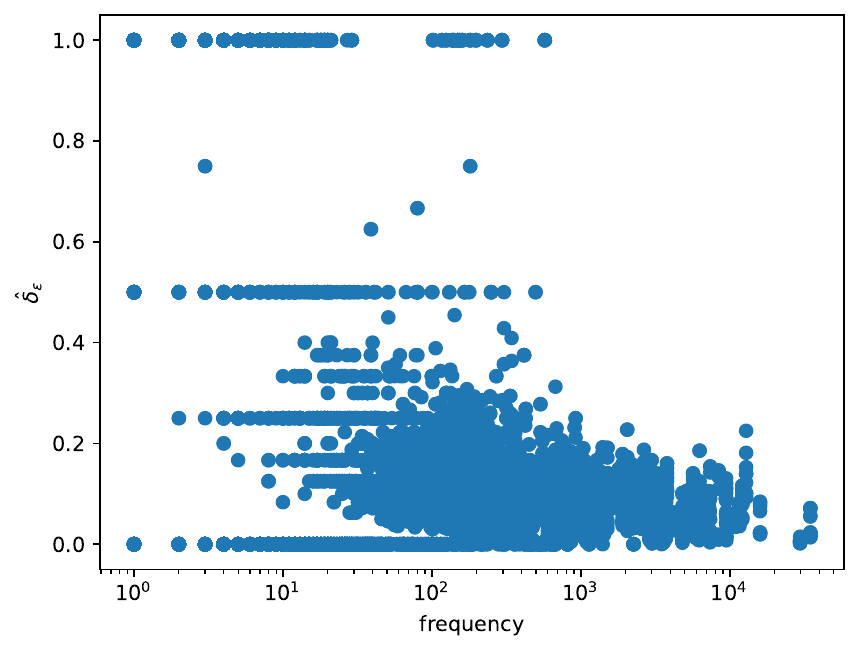}
\caption{This figure demonstrates the weak negative correlation between relation frequency and empirical discrepancy. }\label{fig:relation_delta}
\end{figure}

Figure \ref{fig:relation_alpha} - \ref{fig:relation_delta} demonstrate the relationship between relation frequency and empirical ambiguity/discrepancy.

\section{AI Assistants In Writing}
We use ChatGPT \cite{openai2024chatgpt} to enhance our writing skills, abstaining from its use in research and coding endeavors.

\end{document}